\pdfoutput=1

\documentclass[table,xcdraw,11pt]{article} 

\usepackage[]{acl}

\usepackage{times}
\usepackage{latexsym}

\usepackage{amsmath,amsfonts,bm}

\def\eqref#1{equation~\ref{#1}}

\def\1{\bm{1}}

\DeclareMathAlphabet{\mathsfit}{\encodingdefault}{\sfdefault}{m}{sl}
\SetMathAlphabet{\mathsfit}{bold}{\encodingdefault}{\sfdefault}{bx}{n}

\def\sD{{\mathbb{D}}}

\def\sT{{\mathbb{T}}}

\usepackage{lipsum, babel}
\usepackage[T1]{fontenc}

\usepackage[utf8]{inputenc}

\usepackage{microtype}
\usepackage{float}
\usepackage{booktabs}
\usepackage{graphicx}
\usepackage{subcaption}
\usepackage{xcolor}
\usepackage{footnote}

\usepackage{todonotes}
\usepackage{mathtools}
\usepackage{algorithmicx}
\usepackage{algpseudocode}
\usepackage{amsmath,amssymb}
\usepackage{multirow}
\usepackage{balance}

\usepackage{arabtex}
\usepackage{utf8}
\setcode{utf8}

\makesavenoteenv{tabular}
\makesavenoteenv{table}

\newcommand{\zz}{\mathbf{z}}

\newcommand{\modelM}{\mathbb{M}}

\renewcommand\cite{\citep}

\newcommand{\squeezeupless}{\vspace{-2mm}}
\newcommand{\arastr}[1]{{\small \<#1>}} %

\title{Post-hoc analysis of Arabic transformer models}

\author{
 Ahmed Abdelali$^{\diamondsuit}$ \hspace{2mm} Nadir Durrani$^{\diamondsuit}$ \hspace{2mm} Fahim Dalvi$^{\diamondsuit}$ \hspace{2mm} \hspace{2mm} Hassan Sajjad$^{\clubsuit}$\thanks{\hspace{1.5mm}The work was done while the author was at QCRI}    \\
 $^{\diamondsuit}$Qatar Computing Research Institute, Hamad Bin Khalifa University, Qatar \\
 $^\clubsuit$Faculty of Computer Science, Dalhousie University, Canada \\
 {\{aabdelali, ndurrani,faimaduddin\}@hbku.edu.qa}, \hspace{1mm} hsajjad@dal.ca
}

\begin{document}
\maketitle
\begin{abstract}

Arabic is a Semitic language which is widely spoken with many dialects. Given the success of pre-trained language models, many transformer models trained on Arabic and its dialects have surfaced. While there have been an extrinsic evaluation of these models with respect to downstream NLP tasks, no work has been carried out to analyze and compare their internal representations. 
We probe how linguistic information is encoded in 
the transformer models, trained on different 
Arabic 
dialects. We perform a layer and neuron analysis on the models 
using 
morphological tagging tasks 
for different dialects of Arabic and 
a dialectal identification task. Our analysis enlightens interesting findings such as: 
i) word morphology is learned at the lower and middle layers, ii) while syntactic dependencies are predominantly captured at the higher layers, 
iii) despite a large overlap in their vocabulary, the MSA-based models fail to capture the nuances of Arabic dialects, iv) we found that neurons in embedding layers are polysemous in nature, while the neurons in middle layers are exclusive to specific properties. 

\end{list}
\end{abstract}

\section{Introduction}
\begin{figure}[]
    \centering
    \includegraphics[width=0.75\linewidth]{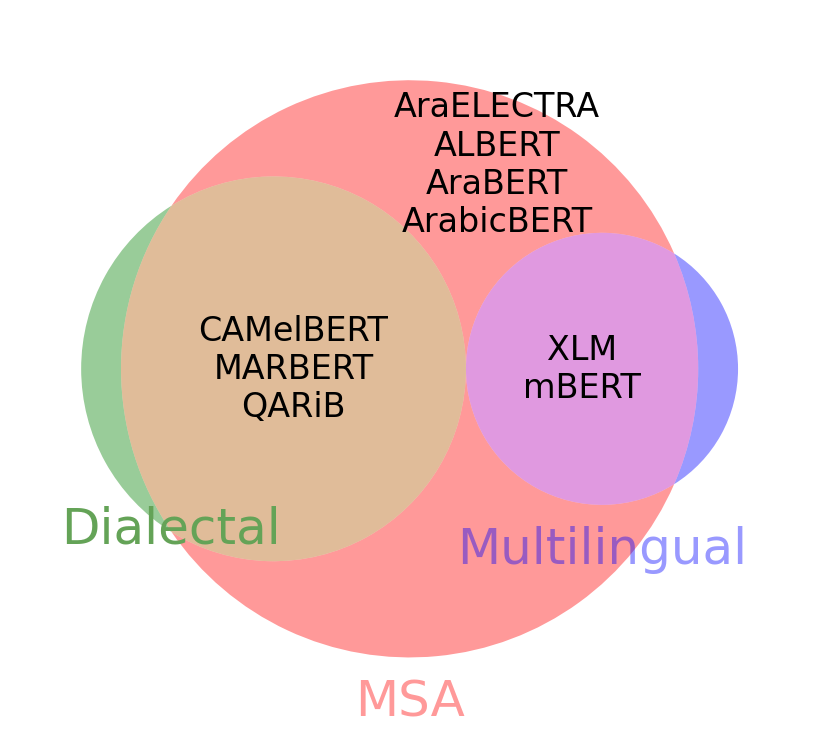}
    \caption{Data regimes of various pre-trained Transformer models of Arabic}
    \label{fig:model-data-venn}
\end{figure}
Arabic is a linguistically rich language, with its structures realized using both concatenative and templatic morphology.
The agglutinating aspect of the language adds to the complexity 
where a given word could be 
formed using multiple morphemes. For example, the word \arastr{fأsqynAkmwh} (fOsqynAkmwh\footnote{Using Safe Buckwalter Arabic (SBA) encoding.} -- and we gave it to you to drink)
combines a conjunction, a verb, and three pronouns. At another longitude, Arabic has three variants: Classical Arabic (CA), Modern Standard Arabic (MSA) and Dialectal Arabic (DA). While the MSA 
is 
traditionally considered as the de facto standard in the written medium and DA being the predominantly spoken counterpart, this has changed recently~\cite{mubarak-darwish-2014-using,
zaidan-callison-burch-2014-arabic, DurraniAI14}. 
Due to the recent influx of Social Media 
platforms, 
dialectal Arabic 
also enjoys a significant presence in the written medium. 

Transfer learning 
using contextualized representations 
in pre-trained language models have revolutionized the arena of downstream NLP tasks. A plethora of transformer-based language models, trained in dozens of languages are uploaded every day now. 
Arabic is no different. Several researchers have released and benchmarked pre-trained Arabic transformer models such as AraBERT~\cite{antoun2020arabert},  ArabicBERT~\cite{safaya-etal-2020-kuisail}, CAMeLBERT ~\cite{inoue-etal-2021-interplay}, MARBERT~\cite{abdulmageed2020arbert} and QARIB~\cite{abdelali2021pretraining} etc. 
These models have 
demonstrated 
state-of-the-art performance on many tasks as well 
as their ability to learn salient features for Arabic. 
One of the main differences 
among these models is the 
genre
and amount of Arabic data they are trained on. 
For example, AraBERT was trained 
only on the MSA (Modern Standard Arabic), ArabicBERT 
additionally used DA during training, and CAMelBERT-mix used a combination of all types of Arabic text for training. 
Multilingual models such as mBERT and XLM are mostly trained on Wikipedia and CommonCrawl data which is predominantly MSA~\cite{10.1145/2911451.2914677}. Figure~\ref{fig:model-data-venn} summarizes the training data regimes of these models.

This large variety of Arabic pre-trained models motivates us to question 
\textbf{how their representations encode 
various linguistic concepts?} To this end, \textbf{we present the first work on interpreting deep Arabic models.} We experiment with nine transformer models including:
five Arabic BERT models, Arabic ALBERT, Arabic Electra, and two multilingual models (mBERT and XLM). We analyze their representations 
using MSA and dialectal parts-of-speech tagging 
and dialect identification tasks. 
This allows us to 
compare the representations 
of Arabic 
transformer models using tasks involving different varieties of Arabic dialects. 

We analyze representations of the network at layer-level 
and at neuron-level 
using diagnostic classifier framework \cite{belinkov:2017:acl,hupkes2018visualisation}. The overall idea is to extract feature vectors from the learned representations and train probing classifiers towards understudied auxiliary tasks (of predicting morphology or identifying dialect). 
We additionally use the 
\emph{Linguistic Correlation Analysis} method~\cite{dalvi:2019:AAAI,durrani-etal-2020-analyzing} to identify salient neurons with respect to a downstream task. 
Our results show that: 


\paragraph{Network and Layer Analysis}
\begin{itemize}
    \item 
    Lower and middle layers 
    capture word morphology 
    \item The long-range contextual knowledge required to solve the dialectal identification is preserved in the higher layers
\end{itemize}

\paragraph{Neuron Analysis}
\begin{itemize}
    \item The salient neurons with respect to a property are well distributed across the network
    \item First (embedding) and last layers of the models contribute a substantial amount of salient neurons for any downstream task
    \item The neurons of embedding layer 
    layer are polysemous in nature while the neurons of middle layers specializes in specific properties
\end{itemize}

\paragraph{MSA vs. Dialect}
\begin{itemize}
    \item Although
    dialects of Arabic are closely related to MSA, the pre-trained models trained using MSA only do not implicitly learn nuances of dialectal Arabic
\end{itemize}

	

	


\newcommand{\sitem}[1]{
    {\setlength\itemindent{15pt} \item[-] #1}
}

\section{Methodology}
\label{sec:methodology}

Our methodology is based on the class of interpretation methods called as the \emph{Probing Classifiers}. The central idea is 
to extract the activation vectors 
from a pre-trained language model as static features. These activation vectors are then trained towards the task of predicting a property of interest, 
a linguistic task that we would like to probe the representation against. The underlying  
assumption is that if the classifier can predict the property, the representations implicitly encode this information. We train layer \cite{belinkov-etal-2020-linguistic} and neuron probes \cite{neuronJournal} using logistic-regression classifiers.  

Formally, consider a pre-trained neural language model $\mathbf{M}$ with $L$ layers: $\{l_1, l_2, \ldots, l_L\}$. Given a dataset $\sD=\{w_1, w_2, ..., w_N\}$ 
with a corresponding set of linguistic annotations $\sT=\{t_{w_1}, t_{w_2}, ..., t_{w_N}\}$, we map each word $w_i$ in the data $\sD$ to a sequence of latent representations: $\sD\xmapsto{\modelM}\zz = \{\zz_1, \dots, \zz_N\}$. The layer-wise probing classifier is trained by minimizing the following loss function:

\begin{equation}
\mathcal{L}(\theta) = -\sum_i \log P_{\theta}(t_{w_i} | w_i) \nonumber
\end{equation}

\noindent where $P_{\theta}(t_{w_i} | w_i) = \frac{\exp (\theta_l \cdot \zz_i)}{\sum_{l'} \exp (\theta_{l'} \cdot \zz_i)} $ is the probability that word $i$ is assigned property $t_{w_i}$. 
 
For neuron analysis, we use \emph{Linguistic Correlation Analysis} (\texttt{LCA}) as described in \cite{dalvi:2019:AAAI}. LCA is also based on the probing classifier paradigm. However, they used elastic-net regularization~\cite{Zou05regularizationand} that enables the selection of both focused and distributed neurons. The loss function is as follows:

\begin{equation}
\mathcal{L}(\theta) = -\sum_i \log P_{\theta}(t_{w_i} | w_i) + \lambda_1 \|\theta\|_1 + \lambda_2 \|\theta\|^2_2 \nonumber
\end{equation}


\noindent The regularization parameters $\lambda_1$ and $\lambda_2$ are tuned using a grid-search algorithm.
The classifier assigns weight to each feature (neuron) which serves as their importance with respect to a class like Noun. We ranked the neurons based on the absolute weights for every class. We select salient neurons for the task such as POS by iteratively selecting top neurons of every class.

A minimum set of neurons is 
identified 
by iteratively selecting top neurons that achieves classification performance comparable (within a certain threshold) 
to the \emph{Oracle} 
%
 -- accuracy of the classifier trained using all the features in the network.

\begin{table}[ht]
\small
\centering
\scalebox{0.9}{
\begin{tabular}{lcrcc}
\toprule \textbf{Data} & \textbf{Size} & \textbf{Tokens} & \textbf{Vocab} & \textbf{Type} \\ 
\midrule
AraBERT & 23GB & 2.7B  & 64K & MSA\\
ArabicBERT & 95GB & 8.2B& 32K  & MSA\\
CAMeLBERT & 167B & 17.3B & 30K & MSA/CA/DA \\
MARBERT & 128GB & 15.6B & 100K & MSA/DA \\
mBERT & - & 1.5B & 110K & MSA\\
QARiB & 127GB & 14.0B & 64K & MSA/DA\\
AraELECTRA & 77GB & 8.6B & 64K & MSA \\
ALBERT & -  & 4.4B & 30K & MSA\\
XLM & 2.5TB & - & 250K & MSA \\

\bottomrule
\end{tabular}
}
\caption{\label{data-stats} Pretrained Models data and statistics.
}
\end{table}


\section{Experimental Setup}

In this section, we describe our experimental setup including the Arabic transformer models, probing tasks that we have used to carry the analysis and the classifier settings.


\subsection{Pre-trained Models}
\label{pretrainedmodels}

We select a number of Arabic transformer models, trained using various varieties of Arabic and based on different architectures. Table~\ref{data-stats} provides a summary of these models. In the following, we describe each model and the dataset used for their training. 

\paragraph{AraBERT} was trained using a combination of 70 million sentences from Arabic Wikipedia Dumps,  1.5B words Arabic Corpus~\cite{elkhair201615} and the Open Source International Arabic News Corpus (OSIAN) from \citep{zeroual-etal-2019-osian}. The final corpus 
contained 
mostly MSA news from different Arab regions.

\paragraph{ArabicBERT}
\newcite{safaya-etal-2020-kuisail}
pretrained a BERT model using a concatenation of Arabic version of OSCAR~\cite{ortizsuarez:hal-02148693}, a filtered subset from Common Crawl and a dump of Arabic Wikipedia totalling to 8.2B words. 

\paragraph{CAMeLBERT}
\newcite{inoue-etal-2021-interplay} combined a 
mixed collection 
of MSA, Dialectal and Classical Arabic texts with a total of 
17.3B tokens. They used the data to pre-train CAMeLBERT-Mix model.  

\paragraph{MARBERT}
\newcite{abdulmageed2020arbert} combined a dataset of 1B tweets that 
covering mostly Arabic dialects and Arabic Gigaword 5th Edition,\footnote{LDC Catalogue LDC2011T11} OSCAR~\cite{ortizsuarez:hal-02148693}, OSIAN~\cite{zeroual-etal-2019-osian} and Wikipedia dump totally up to 15.6B tokens. 

\paragraph{QARIB} 
\newcite{abdelali2021pretraining} combined Arabic Gigaword Fourth Edition,\footnote{LDC Catalogue LDC2009T30} 1.5B words Arabic Corpus~\cite{elkhair201615}, the Arabic part of Open Subtitles~\cite{lison-tiedemann-2016-opensubtitles2016} and 440M tweets 
collected between 2012 and 2020. The data was processed using Farasa \cite{abdelali-etal-2016-farasa}.

\paragraph{ALBERT} used a subset of OSCAR~\cite{ortizsuarez:hal-02148693} and a dump of Wikipedia, selecting around 4.4 Billion words~\citep{ali_safaya_2020_4718724}.
The model differs from 
BERT 
using factorized embedding and 
repeating layers which results in a small memory footprint~\cite{lan2020albert}.

\paragraph{AraELECTRA}  
ELECTRA, 
model \newcite{clark2020electra} is trained to distinguish "real" vs "fake" input tokens generated by another neural network. The Arabic ELECTRA was trained on 77GB of data 
combining OSCAR dataset, Arabic Wikipedia dump, the 1.5B words Arabic Corpus, the OSIAN Corpus and Assafir news articles~\cite{antoun2021araelectra}. 
Different than other models, AraELECTRA uses a hidden layer size of 256 while all other models have 768 neurons per layer. 

\paragraph{Multilingual BERT}
Google research released BERT multilingual base model pretrained on the concatenation of monolingual Wikipedia corpora from 104 languages with a shared word piece vocabulary of 110K. 

\paragraph{XLM} 
\newcite{conneau2020unsupervised} is a multi-lingual version of RoBERTa, trained on 2.5TB CommonCrawl data.  The model is trained on 100 different languages. 

\begin{table*}[!htb]
\centering
\scalebox{0.83}{
\begin{tabular}{l|c|c}
\hline
\multicolumn{1}{c|}{\multirow{4}{*}{ATB}} & Text   & \arastr{واعرب بيرسول عن دهشته ازاء تطور مستواه المتواصل ،} \\
\multicolumn{1}{c|}{}                     & Labels & VBD NNP IN NN NN NN NN DT+JJ PUNC \\
\multicolumn{1}{c|}{}                     & SBA   & wAErb byrswl En dhcth AzAC tTwr mstwAh AlmtwASl , \\
\multicolumn{1}{c|}{}                     & Gloss. & And Peirsol expressed his surprise at the continuous development of his level , \\ \hline
\multirow{4}{*}{CRS}                     & Text   & \arastr{نط أخوي الصغير وجبلي مي} ! \\
                                         & Labels & VERB NOUN+PRON DET+ADJC PREP+VERB+PREP+PRON NOUN PUNC \\
                                         & SBA   & nT Oxwy AlSgyr wjbly my ! \\
                                         & Gloss. & My little brother jumped and brought me water ! \\ \hline
\multirow{4}{*}{DIA}                     & Text   & \arastr{في ناس مناح ما في متلن ، و في ناس منيح اللي ما في متلن} \\
                                         & Labels & ADV NOUN ADJ PART ADV NOUN PUNC CONJ ADV NOUN ADJ PART PART ADV NOUN \\
                                         & SBA   & fy nAs mnAH mA fy mtln , w fy nAs mnyH Ally mA fy mtln \\
                                         & Gloss. & There are good people who are unparalleled, and there are people that it is good they are unparalleled. \\ \hline
\multirow{4}{*}{DID}                     & Text   & \arastr{له مجددا أقول مفقوسة أوي} ! \\
                                         & Labels & lang1 lang1 ambiguous lang2 lang2 other \\
                                         & SBA   & lh mjddA Oqwl mfqwsp Owy \\
                                         & Gloss. & For him again I say (I am) very upset ! \\ \hline
\multirow{4}{*}{GMR}                     & Text   & \arastr{سيف : انا ما غلطت قلت صدق } \\
                                         & Labels & NOUN\_PROP:MS PUNC:- PRON:1S PART:- VERB:P1S VERB:P1S NOUN:MS \\
                                         & SBA   & syf : AnA mA glTt qlt Sdq \\
                                         & Gloss. & Saif: I wasn't wrong, I said the truth. \\
                                         \hline
\end{tabular}
}
      \caption{Examples of Arabic annotated text and their corresponding labels for each task.} 
\label{tasks-exp}
\end{table*}

\subsection{Probing Tasks}

We consider 
morphological tagging on a variety of Arabic dialects and dialectal identification tasks to analyze and compare the models. Below we describe the task details.

\textbf{POS Tagging on Arabic Treebank (ATB)}: The Arabic Treebank Part1 v2.0 and Part3 v1.0 with a total of 515k tokens labeled at the segment level with POS tags. The data is a combination of  newswire  text  from  An-Nahar and Agence France Presse corpus~\cite{Maamouri2004ThePA}. The data is labeled with 42 distinct tags.


\textbf{Gumar POS Tagging on Gulf Arabic (GMR)}: ~\citeauthor{khalifa-etal-2020-morphological}(~\citeyear{khalifa-etal-2020-morphological}) compiled a collection of 15,225 sentences from eight different novels written in the Emirati Arabic dialect from the Gumar Corpus~\cite{khalifa-etal-2018-morphologically}. The data was manually annotated for tokenization, part-of-speech, lemmatization, spelling adjustment, English glosses and sentence level dialect identification,  
using 169 tags.

\textbf{Curras POS Tagging on  Palestinian Arabic dialect (CRS)}: ~\citeauthor{JarrarHAAZ2017Curras}(\citeyear{JarrarHAAZ2017Curras}) collected  around 5K sentences written in Palestinian Arabic dialect from web blogs, Twitter and Facebook comments and transcripts from a TV Shows Watan Aa Watar. The sentences were manually annotated  for part-of-speech (POS), stem, prefix, suffix, lemma, and gloss using 260 tags.

\textbf{POS Multidialects (DIA)}: A total of 1.4k tweets from four Arabic dialects, namely Egyptian (EGY), Levantine (LEV), Gulf (GLF), and Maghrebi (MGR). The tweets were morphologically tagged~\cite{samih-etal-2017-learning} using a reduced subset of 22 tags.

\textbf{Dialect IDentification (DID)}: This task is related to code switching and language identification (LID) between MSA and Egyptian dialect on social media content. 
The data comprises intrasentential code switched sentences (mixing languages between utterances) used for the Second Shared Task on Language Identification in Code-Switched Data. The data contains over 11k sentences, where each token in the sentences is labeled with one of the eight labels:lang1, lang2, fw, mixed, unk, ambiguous, other and named entities (ne)~\cite{Molina2016Overview}.

Figure\ref{tasks-exp} shows examples for each of the probing tasks with their respective labels.

    



\begin{table*}[!htb]
\centering
\scalebox{0.83}{
\begin{tabular}{l|cc|cc|cc|cc|cc|c}
\hline
Task & \multicolumn{2}{c|}{ATB} & \multicolumn{2}{c|}{CRS} & \multicolumn{2}{c|}{DIA} & \multicolumn{2}{c|}{DID} & \multicolumn{2}{c|}{GMR} &  \multirow{2}{*}{Avg. Acc.} \\ \cline{0-10}
Model      & Acc. & Sel. & Acc. & Sel. & Acc. & Sel. & Acc. & Sel. & Acc. & Sel. &  \\ \hline
AraBERT    & 93.9 & 48.1 & 77.3 & 22.1 & 79.0 & 58.1 & 84.7 & 37.4 & 90.4 & 06.4  & 85.06 \\
ArabicBERT & 95.2 & 48.2 & 80.5 & 24.7 & 83.6 & 50.4 & 91.2 & 34.7 & 91.6 & 05.1 & 88.43 \\
CAMelBERT  & \textbf{95.8} & 39.2 & 82.9 & 23.6 & \textbf{86.0} & 37.2 & 92.0 & 21.3 & 93.0 & 05.6 & 89.94\\
MARBERT    & 95.6 & 51.4 & \textbf{84.2} & 27.5 & 84.8 & 48.8 & 93.1 & 33.9 & \textbf{93.4} & 07.3 & 90.22\\
QARiB      & \textbf{95.8} & 50.6 & 84.0 & 28.9 & 85.4 & 45.0 & \textbf{93.3} & 28.8 & 93.3 & 06.7 & \textbf{90.38} \\
mBERT      & 94.4 & 48.8 & 73.7 & 22.7 & 77.6 & 58.4 & 81.7 & 36.3 & 88.0 & 04.4 & 83.08 \\
AraELECTRA & 94.4 & 46.9 & 72.7 & 28.4 & 79.0 & 56.2 & 87.9 & 34.3 & 89.1 & 08.1 & 84.64\\
ALBERT     & 95.2 & 40.9 & 77.0 & 28.3 & 82.1 & 39.8 & 88.3 & 27.1 & 90.2 & 09.4 & 86.56\\
XLM & 95.7 & 43.7 & 75.0 & 20.6 & 78.9 & 42.1 & 86.7 & 29.3 & 88.2 & 06.0 & 84.90 \\ \hline
\end{tabular}
}
      \caption{Classifier performance on Test sets using  \textbf{top layers}} 
\label{models-accur}
\end{table*}

\begin{table*}[!htb]
\centering
\scalebox{0.83}{
\begin{tabular}{l|cc|cc|cc|cc|cc}
\hline
Task       & \multicolumn{2}{c|}{ATB} & \multicolumn{2}{c|}{CRS}  & \multicolumn{2}{c|}{DIA}  & \multicolumn{2}{c|}{DID} & \multicolumn{2}{c}{GMR}  \\ \hline
Threshold $\delta$ & \multicolumn{2}{c|}{5\%} & \multicolumn{2}{c|}{10\%} & \multicolumn{2}{c|}{10\%} & \multicolumn{2}{c|}{7\%} & \multicolumn{2}{c}{5\%}  \\ \hline
Model      & Acc. & Sel.   & Acc. & Sel.   & Acc. & Sel.   & Acc. & Sel.   & Acc. & Sel.  \\ \hline
AraBERT & 93.4 & 48.8 & 82.1 & 33.5 & 79.3 & 39.9 & 86.0 & 21.5 &  89.9 & 18.1 \\
ArabicBERT & 94.0 & 50.8 & 83.6 & 31.6 & 83.3 & 44.9 & 90.1 & 26.3 &  91.0 & 12.5 \\
CAMelBERT & 94.9 & 51.1 & 86.1 & 37.0 & 85.1 & 47.5 & 91.0 & 27.2 &  92.6 & 22.6 \\ 
MARBERT & 94.5 & 51.6 & 86.2 & 30.0 & 84.2 & 48.5 & 91.6 & 29.2 &  92.3 & 15.0 \\
QARiB & 95.0 & 52.7 & 86.1 & 31.0 & 83.6 & 46.7 & 91.7 & 30.0 &  92.4 & 11.6 \\
mBERT & 94.1 & 48.5 & 78.4 & 33.1 & 77.5 & 37.7 & 83.2 & 17.0 &  87.6 & 13.4 \\ 
AraELECTRA & 91.2 & 53.3 & 79.0 & 33.9 & 79.4 & 45.0 & 88.2 & 25.6 & 87.6 & 13.1 \\
ALBERT & 94.7 & 56.8 & 80.7 & 33.7 & 81.8 & 47.4 & 88.5 & 25.9 &  89.8 & 12.0 \\
XLM & 95.3 & 51.8 & 78.5 & 27.9 & 79.0 & 44.5 & 86.5 & 21.7 & 88.0 & 12.4 \\ \hline    


\end{tabular}
}
\caption{Classifier performance on Test sets using \textbf{top neurons} as features
\label{tab:neuron-models-accur}}
\end{table*}

\subsection{Post-hoc Classifier}
We used the NeuroX toolkit~\cite{neuroX:aaai19:demo} to perform our analysis. 
Our probe is a linear classifier with categorical cross-entropy loss, optimized by Adam \cite{kingma2014adam}. For neuron-analysis, the classifier additionally used the elastic-net regularization~\cite{Zou05regularizationand}. The regularization weights are trained using grid-search algorithm. Training is run with shuffled mini-batches of size 512 and stopped after 10 epochs.  
Linear classifiers are a popular choice in analyzing deep NLP models due to their better interpretability \cite{qian-qiu-huang:2016:EMNLP2016}. \citet{hewitt-liang-2019-designing} have also shown linear probes to have higher \emph{Selectivity}, a property deemed desirable for more interpretable probes.  We perform control task experiments to
ensure that our probes are reflective of the linguistic knowledge that representations capture.
For sub-word based models, we use the average activation value \cite{durrani-etal-2019-one} to be the representative of the word.
We additionally normalize the embeddings using znorm as it has shown to provide better ranking of neurons with respect to a property \cite{sajjad_static_arxiv}.

\section{Analysis and Discussion}

 
 Our goal is to carry out a comparative investigation of  the knowledge encoded in 
 different Arabic transformer models. 
First we compare the representations in terms of how much linguistic information is preserved in the network using 
the overall accuracy 
on the understudied auxiliary tasks. Then we 
analyze how 
such information is preserved across individual layers of the model. Lastly, we analyze the distribution of neurons across the model 
with respect to these tasks.
 

\subsection{Network Analysis}

We use the feature vectors\footnote{We concatenated the features from all layers of the network to train the classifier.} generated from different dialects of Arabic to train post-hoc classifiers towards the task of predicting morphology in these dialects or predicting the dialect themselves. Table \ref{models-accur} gives accuracy of the classifiers on different dialectal tasks. Firstly, 
the high accuracy numbers show that the representations learn non-trivial linguistic knowledge. We found all the models to do well on the task of predicting MSA morphology unsurprisingly, 
since all these models have been trained on 
a large amount of MSA data. Contrastingly, the performance varied a lot on the dialectal tasks with different models giving optimal performance on different dialects. Note that the models that were trained only using MSA performed much worse despite the fact that MSA and dialect have a significant vocabulary overlap. This shows that \textbf{to capture specific dialectal nuances these transformer models need to train 
on dialectal data.} Comparing the models, we found dialectal models (QARiB, MARBERT and CAMeLBERT) to perform 
considerably well across all the tasks. Lastly the high selectivity numbers in Table \ref{models-accur} validate the fact that our classifiers are not memorizing the tasks and are a true reflection of the knowledge captured within the underlying representations.

\begin{figure*}[ht]
\centering
\begin{subfigure}{0.40\textwidth}
  \centering
  \includegraphics[width=\linewidth]{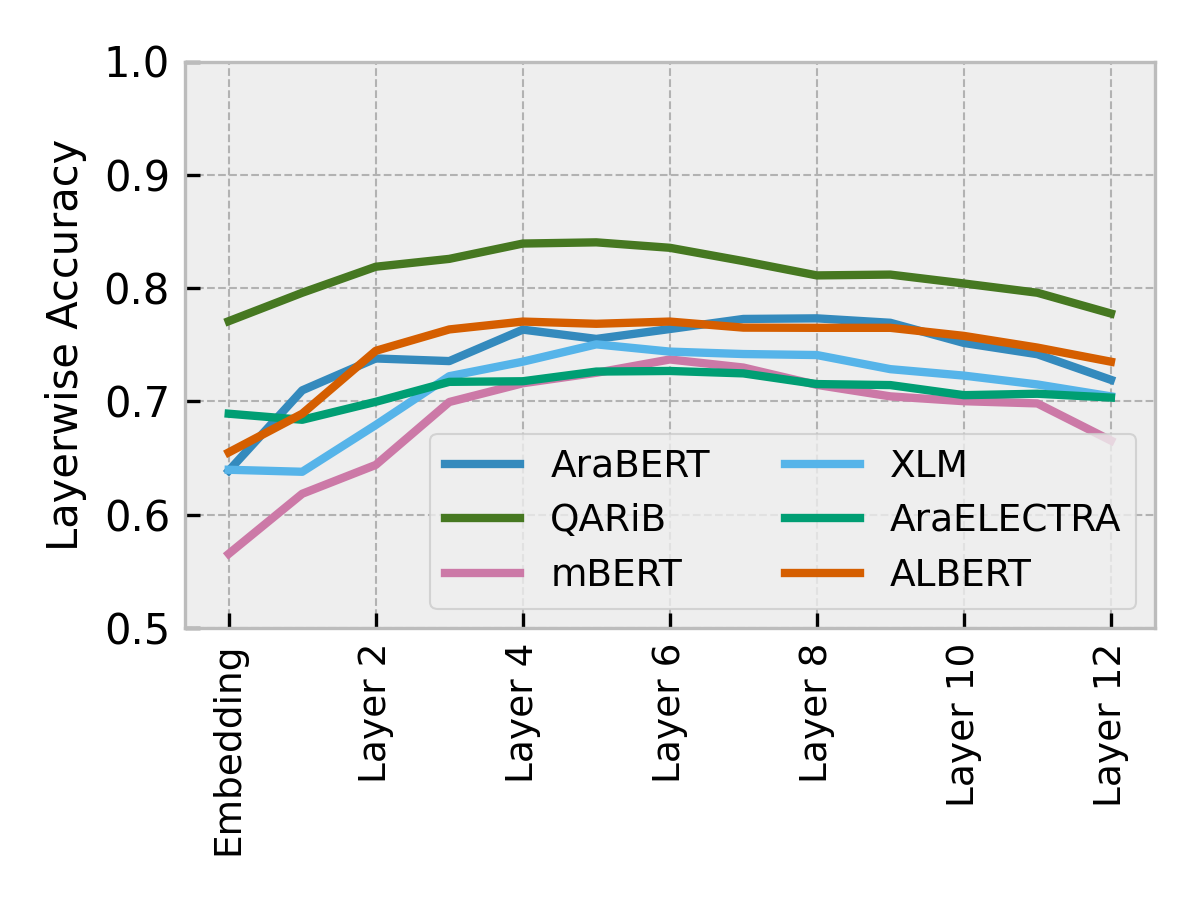}
  \caption{CRS}
  \label{fig:crs_layerwise_acc}
\end{subfigure}
\begin{subfigure}{0.40\linewidth}
  \centering
  \includegraphics[width=\linewidth]{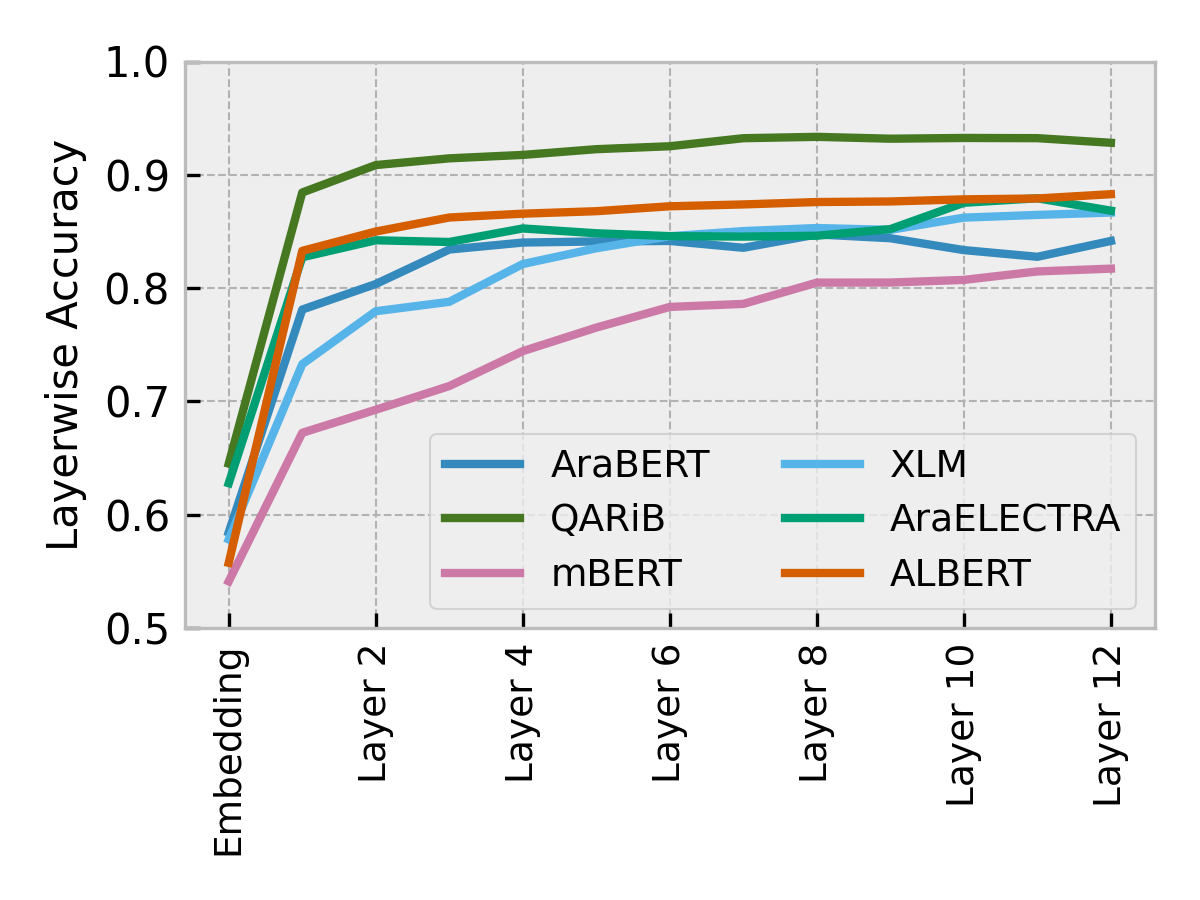}  
  \caption{DID}
  \label{fig:did_layerwise_acc}
\end{subfigure}
\caption{Layer-wise accuracy for different selected tasks.}
\label{fig:layers_accuracy}
\end{figure*}

\subsection{Layer-wise Analysis}
\label{sec:layerWise}

We now analyze how the understudied linguistic knowledge is distributed 
across the layers.  We train a classifier for each probing task using representations of individual layers as features. The performance of the classifier serves as a proxy to the amount of task knowledge learned in each layer representation. 
Figure~\ref{fig:layers_accuracy} provides per-layer accuracy for the CRS (morphological tagging for Palestinian dialect) and DID (Dialect Identification) tasks.\footnote{We limit the presentation to fewer models for clarity purposes. Our observations consistently hold for all dialectal tasks. See Appendix for complete results.}  We found that \textbf{the word morphology is captured predominantly in the lower layers of the model}, retained in the middle layers before declining in the final layers. \textbf{The higher layers are reserved for complex phenomenon such as capturing 
non-local dependencies.} This is confirmed from our DID results. Identifying dialect requires learning non-local dependencies and sentence level phenomenon to accurately predict the dialect. For example, a lexical form can belong to two different dialects depending on the context 
to disambiguate the dialect of the word. For example, \arastr{حاجة} ``HAjp'' (thing or need) is MSA in the context:
\arastr{لست في حاجة لأن} ``lst fy HAjp lOn'' (I am not in \textbf{need} to) 
or Egyptian:
\arastr{مفيش حاجة أصعب من} ``mfyc HAjp OSEb mn'' (there is no \textbf{thing} difficult than). 
The contextual knowledge is essential to disambiguate in such cases. 

\subsection{Neurons Analysis}

We now study how the information is spread across neurons instead of layer by carrying a fine-grained analysis.
We discover neurons 
that learn a particular linguistic property using LCA \cite{dalvi:2019:AAAI} and analyze: i) how many neurons 
can sufficiently capture a concept, ii) how these neurons are distributed across the layers. 
%
LCA provides a ranked list of neurons with respect to the understudied property. We select a minimumal set of top neurons from the ranked list that yield close to the oracle performance.\footnote{Accuracy when using the entire network or best layer, whichever is higher.} 

\paragraph{Minimal Neurons:} 
We found 5\% neurons to be optimal for ATB, 
and GMR; while 10\%  for both CRS and DIA tasks, due to their more granular tag-set. For the DID task, we found 7\% neurons to be optimal (Table \ref{tab:neuron-models-accur} shows results -- 
please also see Appendix for a more detailed result using different neuron thresholds). Our results show that \textbf{a small subset of features can achieve close to oracle performance.} This 
entails that 
re-trainable features are 
available 
in the network as also shown by \newcite{dalvi-etal-2020-analyzing}. 
%
Such a finding 
entails interesting frontier in efficient feature-based transfer learning, which is considered as a viable alternative to the traditional fine-tuning based transfer learning \cite{peters-etal-2019-tune,durrani-etal-2021-transfer,alrowili-shanker-2021-arabictransformer-efficient}. 

\begin{figure*}[ht]
\begin{subfigure}{0.32\textwidth}
  \centering
  \includegraphics[width=\linewidth]{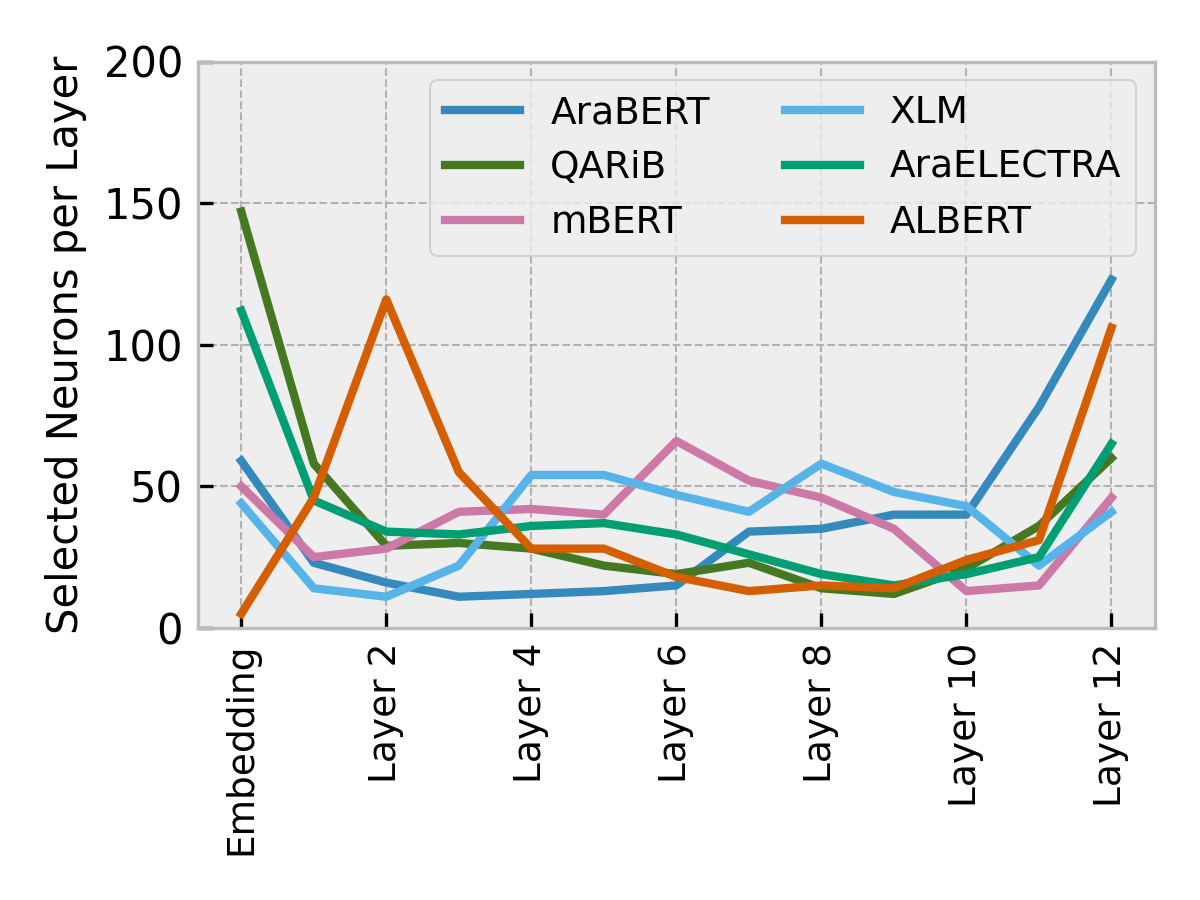}
  \caption{ATB}
  \label{fig:atb_neurons_layer}
\end{subfigure}
\begin{subfigure}{0.32\textwidth}
  \centering
  \includegraphics[width=\linewidth]{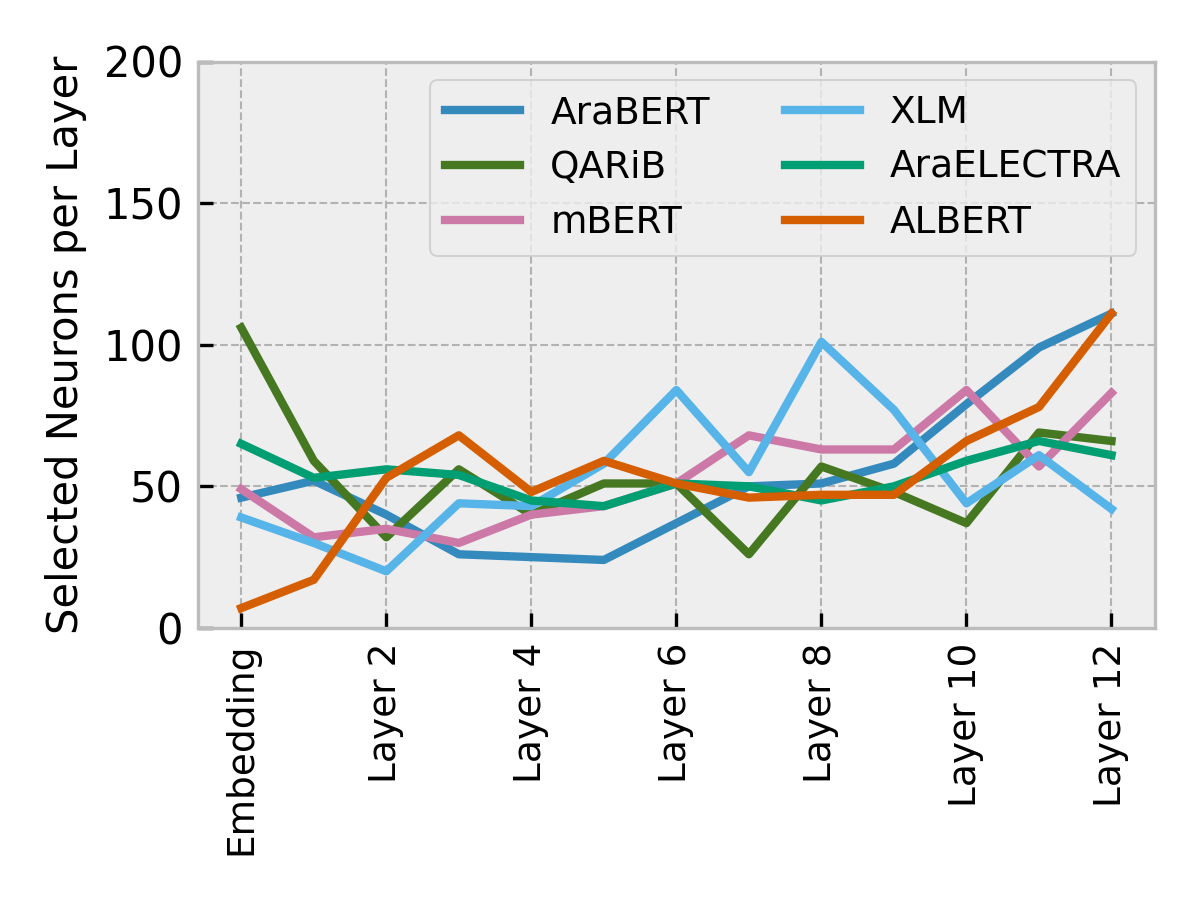}
  \caption{DIA}
  \label{fig:did_neurons_layer}
\end{subfigure}
\centering
\begin{subfigure}{0.32\textwidth}
  \centering
  \includegraphics[width=\linewidth]{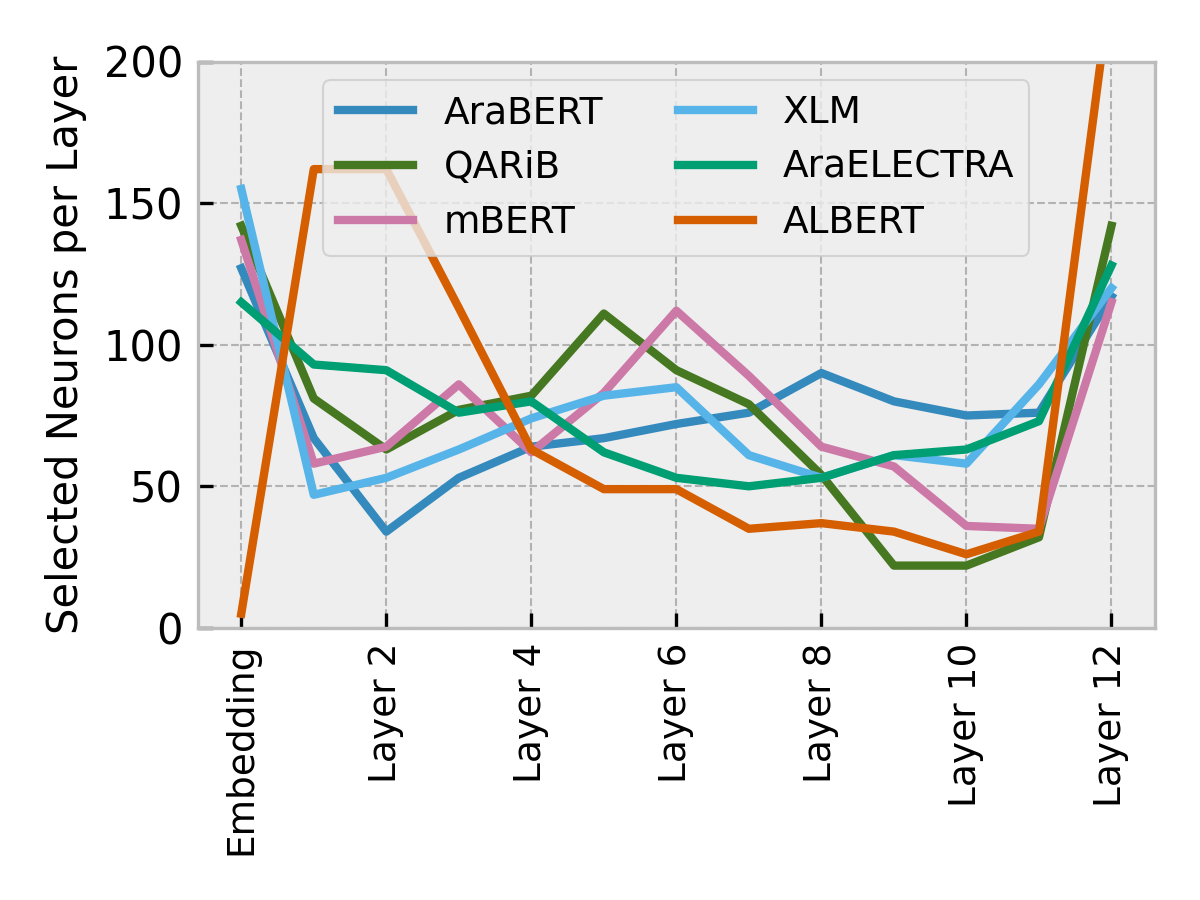}
  \caption{DID}
  \label{fig:atb_neurons_layer}
\end{subfigure}
\caption{Distribution of selected neurons across the layers}
\label{fig:neurons_layer}
\end{figure*}


\paragraph{Neuron Distribution:} Let us now turn our attention towards how these neurons are distributed in the network.
In Figure \ref{fig:neurons_layer} we plot salient neurons across the layers (See Appendix for all the tasks). A dominant pattern that we observed was that the \textbf {embedding and final layers of the model consistently contribute the most number of salient neurons}. This entails that while the neurons in middle layers capture intricate details of the task, the input and output layers of the model that are closer to the actual words possess most lexical information required to for accurate predictions. The model uses the embedding layer to focus on more localized information and final layers to capture contextual dependencies.  An exception to this overall pattern was the ALBERT model, where the embedding layer has close to zero contribution in the salient neurons and relatively higher number of neurons from the initial contextualized layers. Recall that ALBERT has a different architecture where parameters are shared across the encoder layers. Moreover the model factorizes the embedding layer. These architectural choices perhaps explain the difference of neuron distribution pattern. A detailed analysis of word embedding layers using lexical tasks such as word similarity and word relatedness is required to fully understand this. 

\begin{figure*}[ht]
\centering
	\begin{subfigure}{1\textwidth}
		\centering
		\includegraphics[trim=0 0 0 0,clip,width=\linewidth]{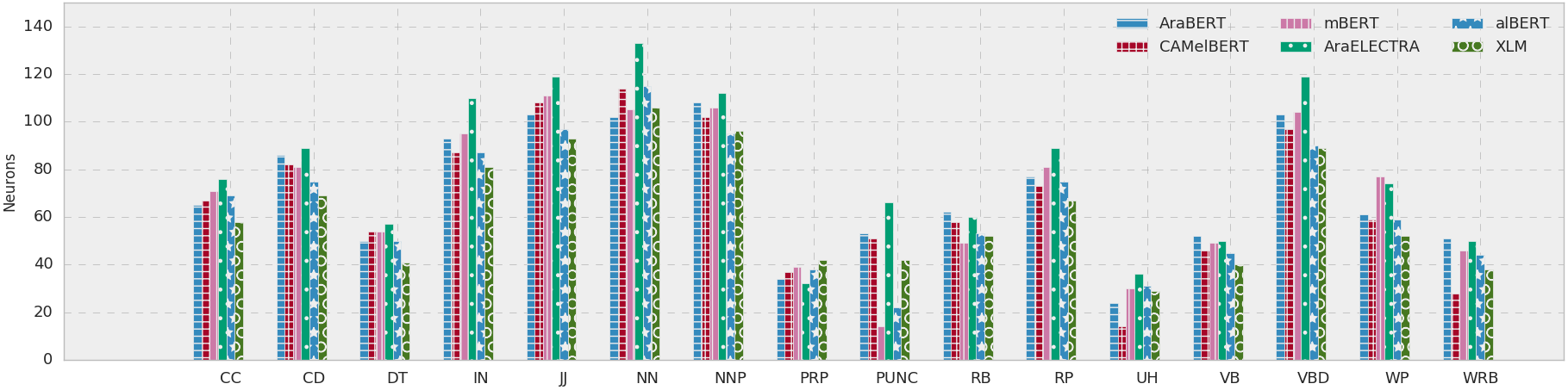}
	\end{subfigure}
	\caption{Distribution of neurons per property (ATB)}
	\label{fig:neurons_per_prop}
\end{figure*}



\paragraph{Property Distribution:}

We have seen how salient neurons distribute across the network. Now we analyze how these neurons distribute across sub-properties within a task. A morphological tagging task for example is composed of different properties such as Noun, Verb, Adjective etc. In Figure \ref{fig:neurons_per_prop} we plot the number of salient neurons required to capture different properties on the task of predicting classes in the ATB task. We observed that \textbf{closed class categories such as personal pronoun (PRP) are localized to fewer neurons, where as the open-class words such as past-tense verbs (VBD) that exhibit a variety of roles in different contexts require a large number of neurons.} We found this observation to be true for all the models across different dialectal tasks (Please see Appendix for more results).

\begin{figure*}[ht]
\begin{subfigure}{.36\textwidth}
  \centering
  \includegraphics[width=\linewidth]{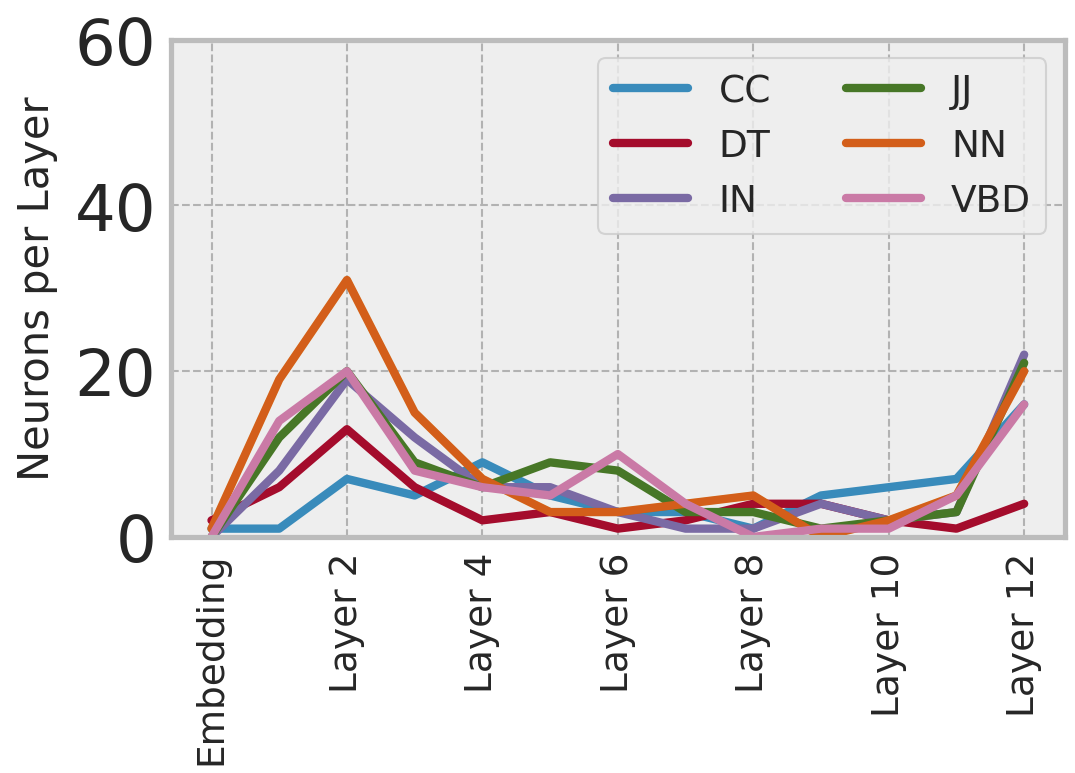}
  \caption{ALBERT}
  \label{fig:pos_atb_albt}
\end{subfigure}
\begin{subfigure}{.31\textwidth}
  \centering
  \includegraphics[width=\linewidth]{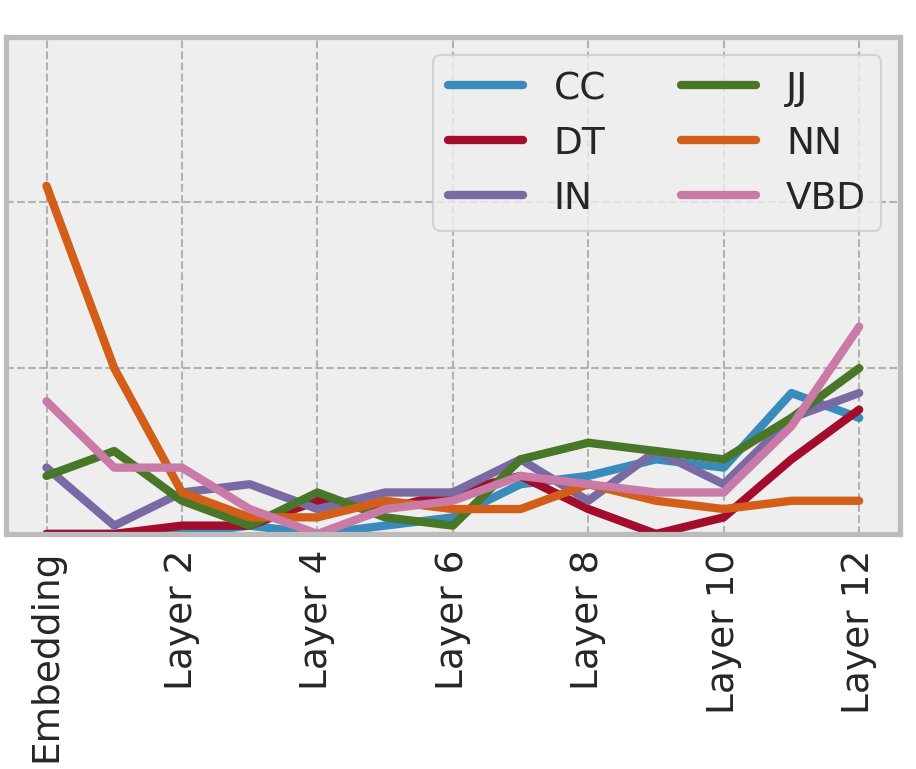}
  \caption{AraBERT}
  \label{fig:pos_atb_arabt}
\end{subfigure}
\begin{subfigure}{.32\textwidth}
  \centering
  \includegraphics[width=\linewidth]{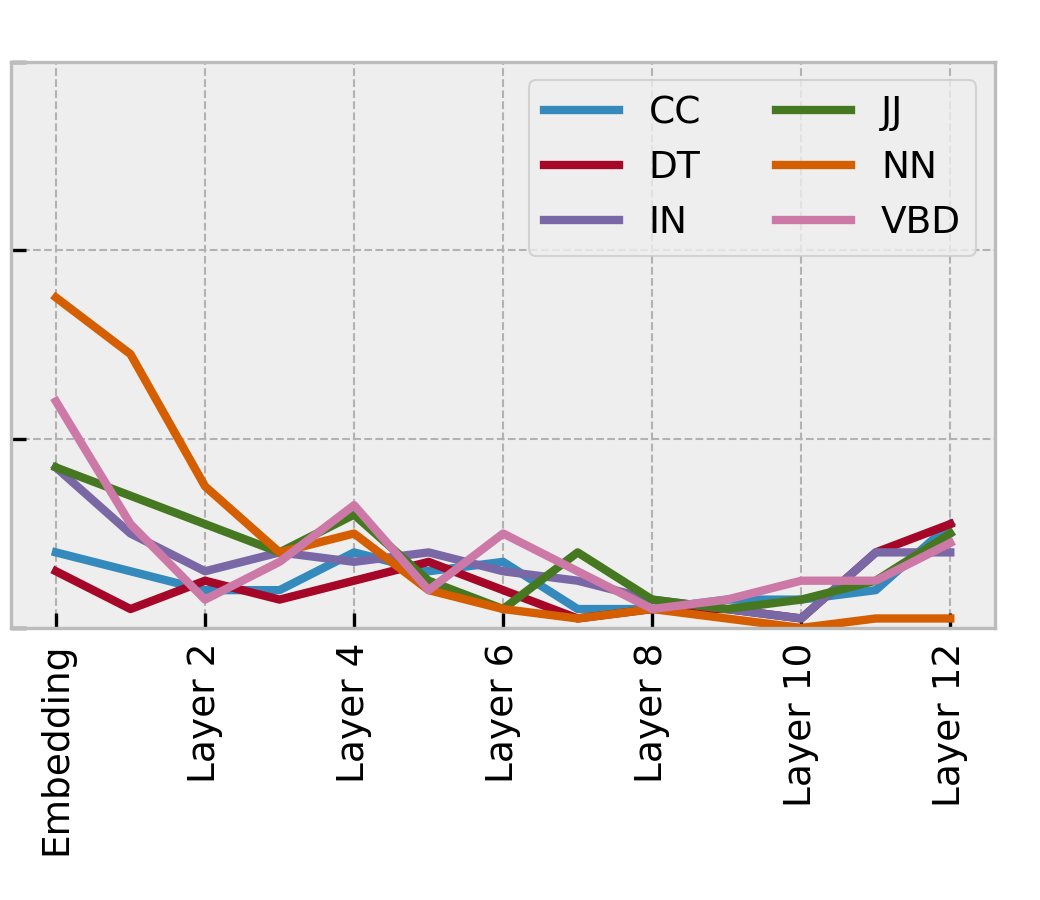}
  \caption{QARiB}
  \label{fig:pos_atb_qarib}
\end{subfigure}
\caption{Property-wise distribution of neurons across the layers in ATB}
\label{fig:pos_propertywise_layer}
\end{figure*}

\paragraph{Layer-wise Property Distribution:} We also analyzed \textbf{how individual properties are encoded across the layers in the network}, Do they have similar neuron distribution pattern or are the specific properties 
learned more on higher layers than lower layers and vice versa? Figure~\ref{fig:pos_propertywise_layer} shows the distribution of selected neurons of ALBERT, AraBERT and QARiB for a few 
properties. We observed a very consistent pattern to the overall neuron distribution that we 
saw in Figure \ref{fig:neurons_layer}. For most of the properties salient neurons were contributed from the embedding and final layers, and middle layers contributed less than 20 neurons. Another interesting pattern to be noted is that noun neurons were more prevalent in the embedding layer (layer 1-2 for ALBERT) but verb neurons were dominantly found in the final layers. Verbs are considered to be structural center in linguistic theories as they connect to all other syntactic units in a sentence~\cite{hudson_2010}. This further reinforces our result that \textbf{the higher layers of the model capture long distance 
dependencies}.

\begin{figure*}[ht!]
\begin{subfigure}{0.33\textwidth}
  \centering
  \includegraphics[width=0.95\linewidth]{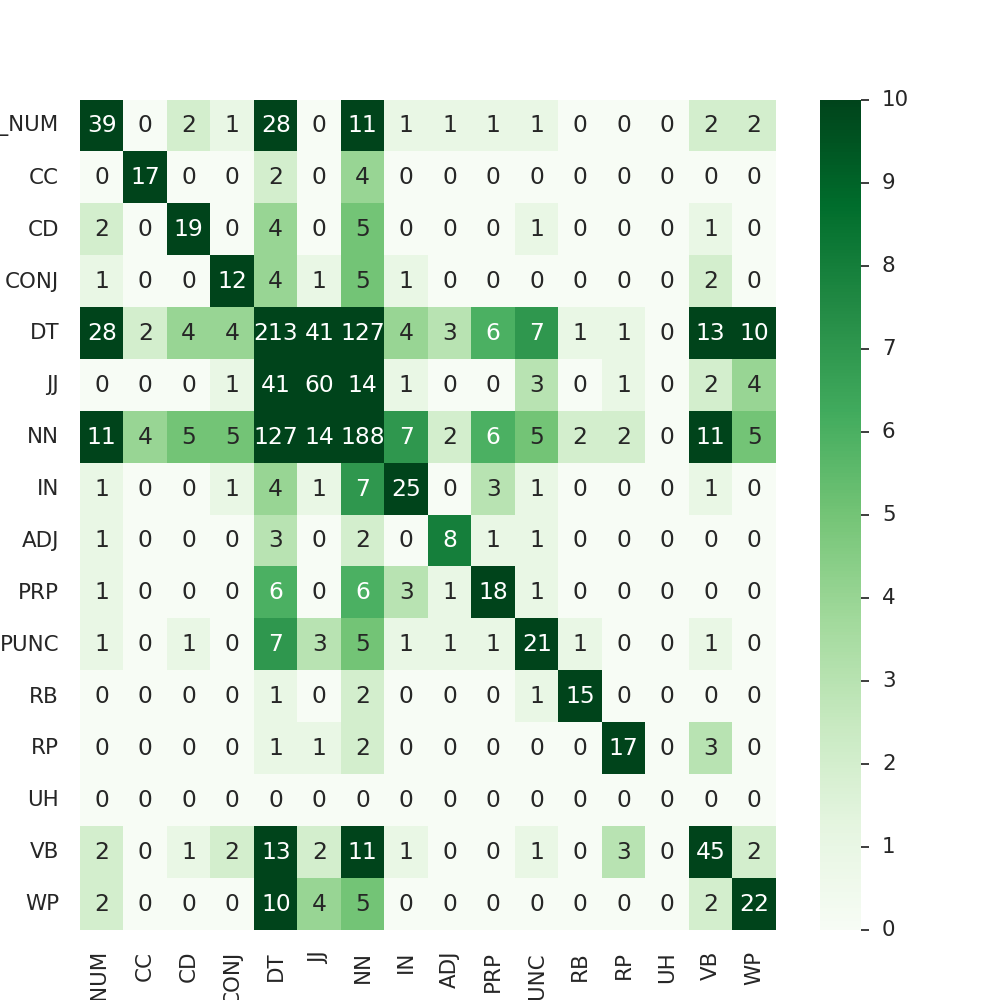}  
  \caption{Layer 0 }
  \label{fig:layers_overlap_0_atb}
\end{subfigure}
\begin{subfigure}{0.33\textwidth}
  \centering
  \includegraphics[width=0.95\linewidth]{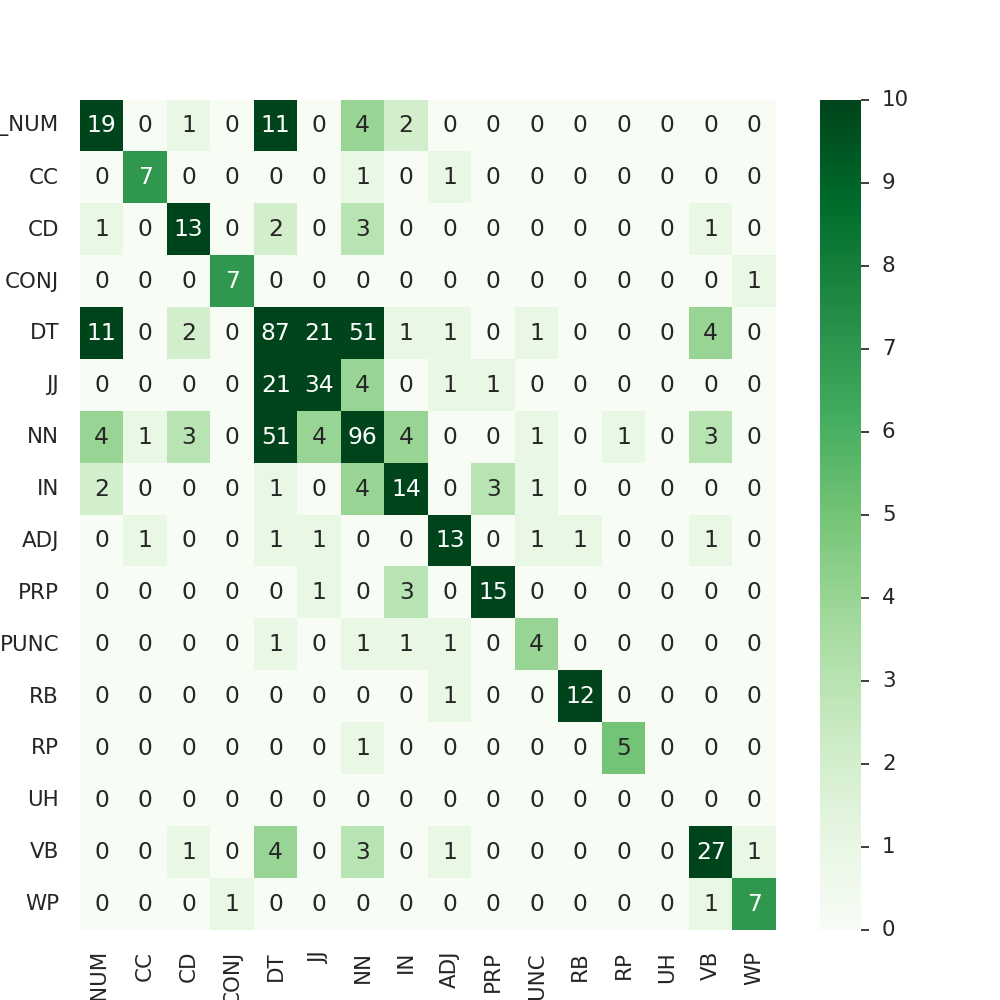}  
  \caption{Layer 1}
  \label{fig:layers_overlap_1_atb}
\end{subfigure}
\begin{subfigure}{0.33\textwidth}
  \centering
  \includegraphics[width=0.95\linewidth]{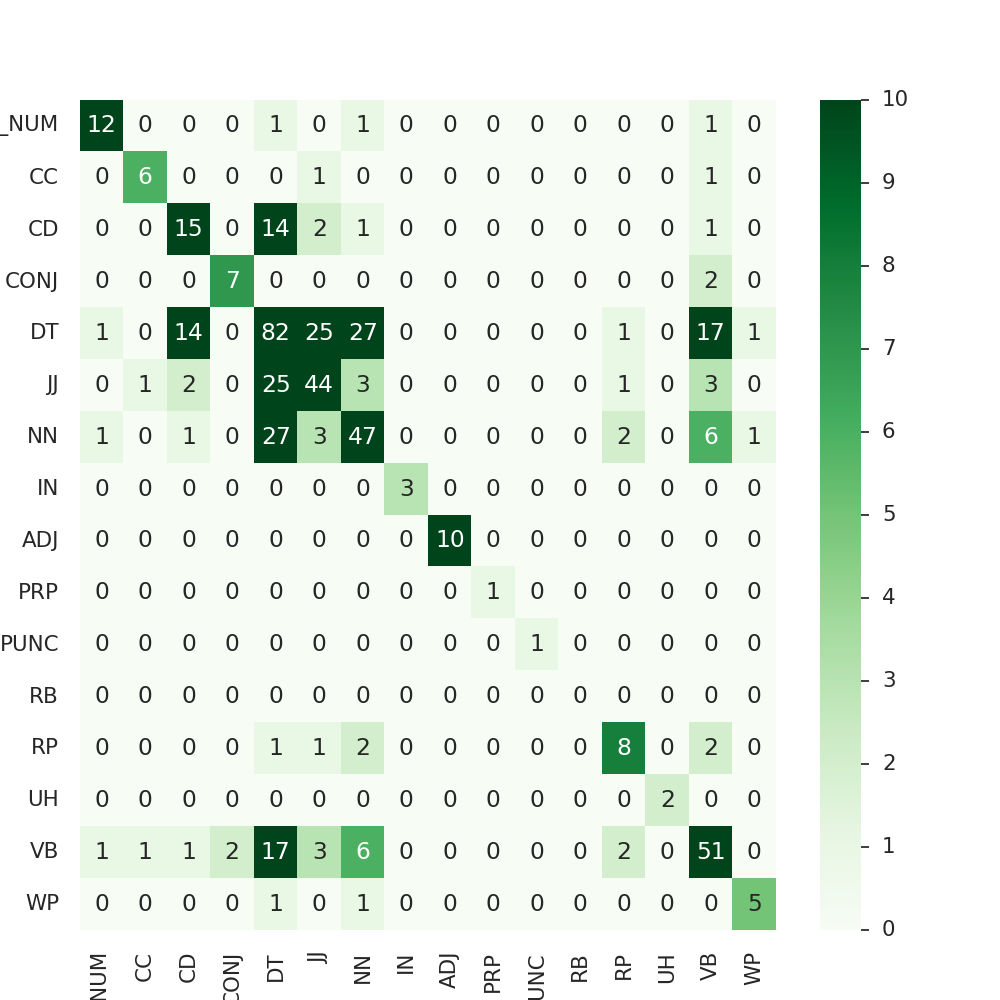}  
  \caption{Layer 5}
  \label{fig:layers_overlap_5_atb}
\end{subfigure}
\begin{subfigure}{0.33\textwidth}
  \centering
  \includegraphics[width=0.95\linewidth]{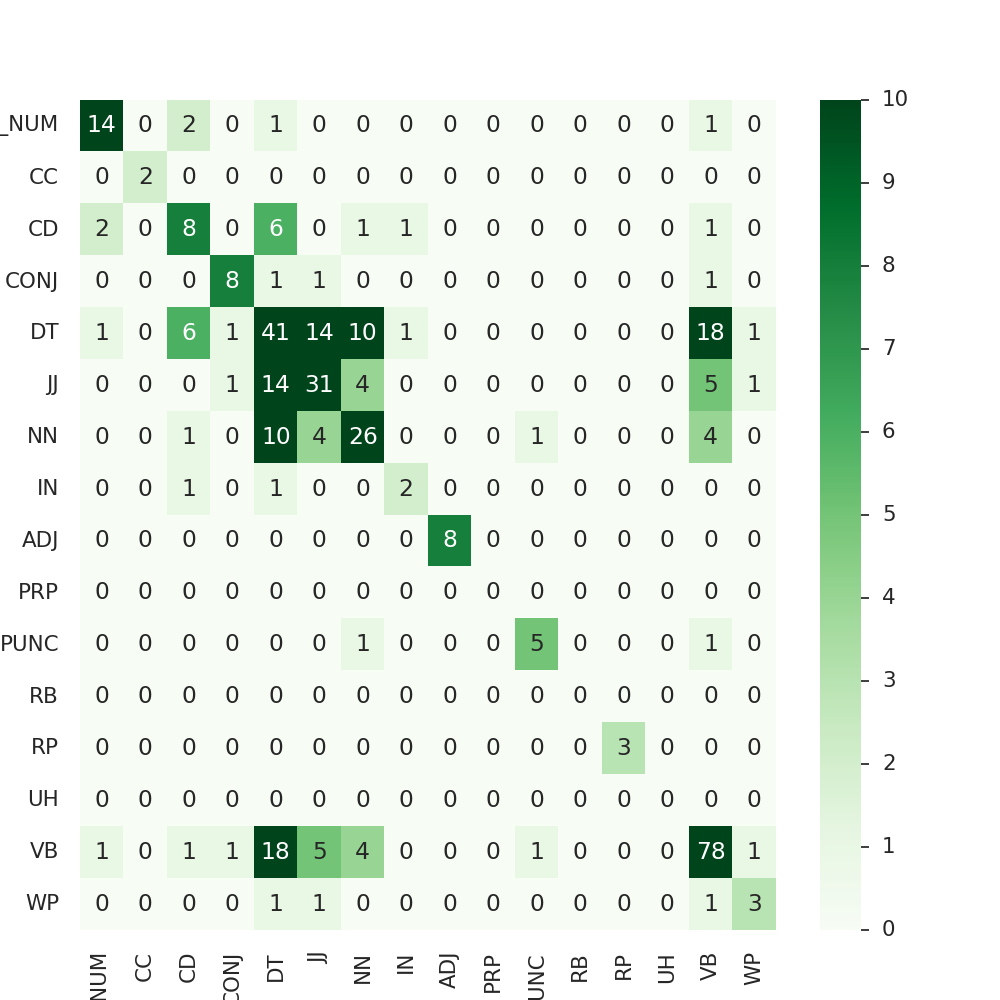}  
  \caption{Layer 6}
  \label{fig:layers_overlap_6_atb}
\end{subfigure}
\begin{subfigure}{0.33\textwidth}
  \centering
  \includegraphics[width=0.95\linewidth]{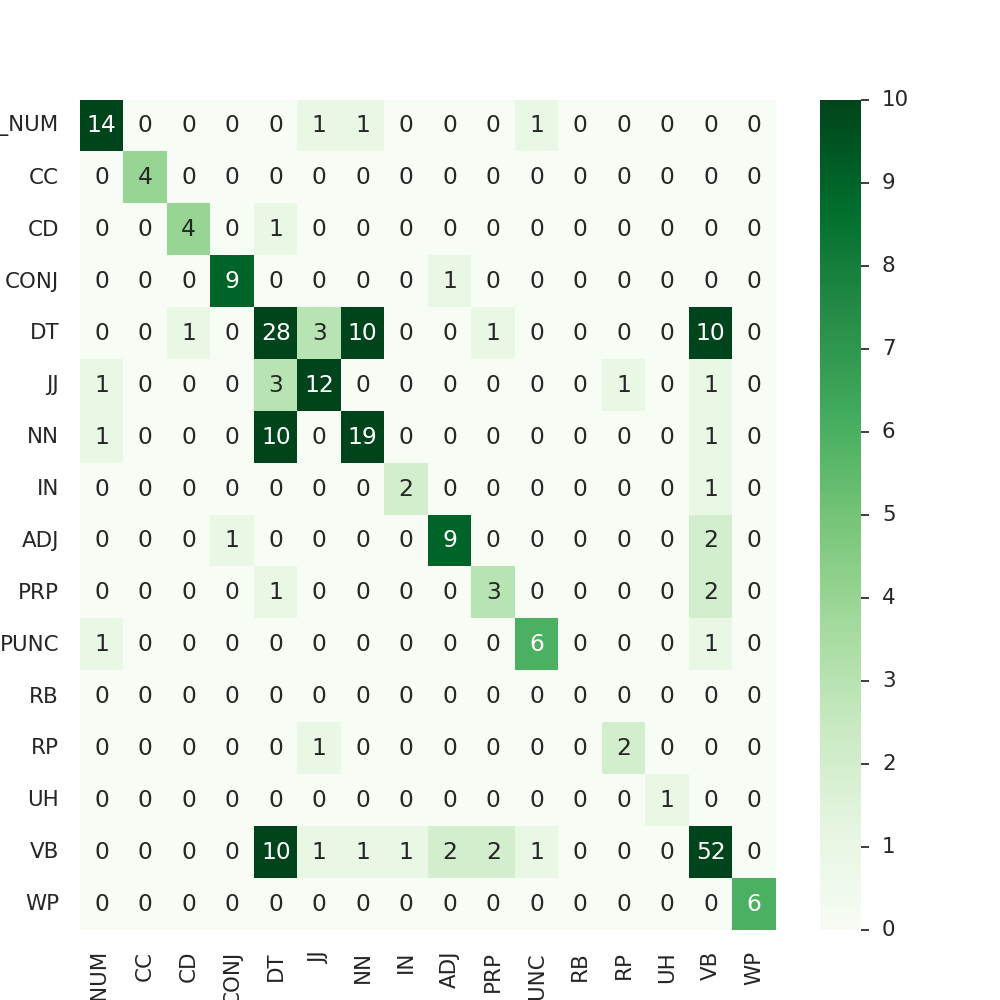}  
  \caption{Layer 11 }
  \label{fig:layers_overlap_11_atb}
\end{subfigure}
\begin{subfigure}{0.33\textwidth}
  \centering
  \includegraphics[width=0.95\linewidth]{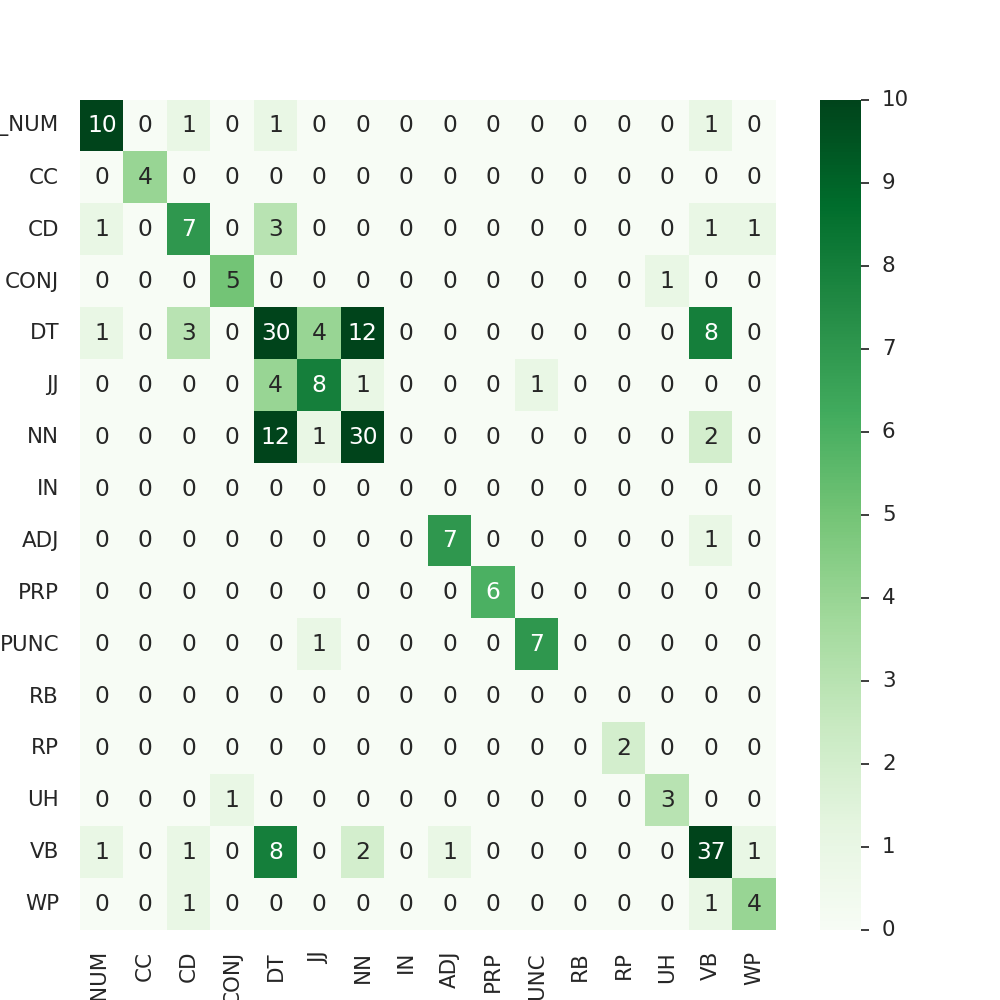}  
  \caption{Layer 12}
  \label{fig:layers_overlap_12_atb}
\end{subfigure}
\caption{QARiB: Neurons overlap across the ATB properties}
\label{fig:layers_overlap}
\end{figure*}


\paragraph{Polysemous Neurons:} Neurons are multi-variate in nature and may capture multiple concepts. For example \newcite{indivdualneuron:arxiv19} discovered switch neurons that activate positively for present-tense verbs and negatively for the past-tense verbs in LSTM encoders. We also analyzed the overlaps between salient neurons that learn different linguistic properties in an attempt to discover polysemous neurons. Figure~\ref{fig:layers_overlap} shows the overlap of neurons across properties in different layers in the QARiB model. The zeros 
means that none of the top neurons between the properties overlap.  Note that there is a high concentration of overlapping neurons between determiners (DT), adjectives (JJ) and nouns (NN) or between determiners and verbs. The 
intersection was around 54\% 
in the case of Determiner ``DT'' and Noun ``NN''. We believe this is an artifact of concatenative morphology that Arabic exhibits, where 
it is common for affixes such as preposition or determiner to join with nouns or adjectives to form composite constructions. We also observed that the number of polysemous neurons exist more dominantly in the embedding layer. Higher layers (Exp. Figure~\ref{fig:layers_overlap_11_atb} and \ref{fig:layers_overlap_12_atb}) show less shared and overlapping neurons.

\section{Related Work}
\label{sec:rw}

Work done on interpreting deep NLP models can be broadly classified into \emph{Concept Analysis} and \emph{Attribution Analysis}. The former thrives on post-hoc decomposability, where we analyze representations to uncover linguistic (and non-linguistic) phenomenon that are captured as the network is trained towards any NLP task \cite{conneau2018you,liu-etal-2019-linguistic,tenney-etal-2019-bert, sajjad-etal-2022-analyzing, dalvi2022discovering} and the latter characterize the role of model components and input features towards a specific prediction \cite{linzen_tacl,gulordava-etal-2018-colorless,marvin-linzen-2018-targeted}. Our work falls into the former category. We carry out a layer and neuron-wise analysis on the Arabic transformer models. We used Diagnostic classifiers \cite{belinkov:2017:acl} to train layer and neuron-wise probes towards predicting linguistic properties of interest. \textbf{To the best of our knowledge this is the first work on analyzing Arabic transformer models.}
\newcite{suau2020finding} used \emph{max-pooling} to identify relevant neurons (aka \emph{Expert units}) in pre-trained models, with respect to a specific concept (for example word-sense). \newcite{Mu-Nips} proposed a \emph{Masked-based Corpus Selection} method to determine important neurons with respect to a concept. See \newcite{neuronSurvey} for a comprehensive survey of these techniques. We used the \emph{Linguistic Correlation Analysis} of \newcite{dalvi:2019:AAAI} to perform neuron analysis.

\section{Conclusion and Future Work}

In this paper we carry out a post-hoc 
analysis on a number of Arabic transformer models using five linguistic tasks. Our results enlighten interesting insights: i) neural networks learn non-trivial amount of linguistic knowledge with lower and middle layers capturing word morphology and higher layers learning more universal phenomenon, ii) we found that salient neurons are distributed across the network, but some layers contribute more salient neurons towards a task, iii) we found some neurons to be polysemous in nature while other capturing very specialized properties, iv) lastly we showed that MSA-based 
models do not capture dialectal nuances despite having a large overlap with dialects.
For future work, we aim to expand this analysis to include more tasks and explore related languages such as 
the families of Semitic, Germanic or Latin languages. 
\bibliography{anthology}
\bibliographystyle{acl_natbib}
\pagebreak
\begingroup
\onecolumn 
\appendix
\section{Appendix}
\label{sec:appendix}

Table~\ref{tab:thresholdselection} shows the performance loss for different thresholds
. Highlighted thresholds were selected based on the 1\% average performance loss. For the case of DIA, some overfitting is noticeable. Such case is reported in literature where the classifiers with large contextualized
vectors tend to overfit when supervised data is insufficient~\cite{hameed}.

\begin{table*}[ht]
\centering
\scalebox{0.83}{
\footnotesize
\begin{tabular}{|c|l|c|c|c|c|c|c|c|c|c|}
\hline
\multicolumn{1}{|l|}{Task} &  Threshold       & \rotatebox{270}{AraBERT} & \rotatebox{270}{ArabicBERT} & \rotatebox{270}{CAMelBERT} & \rotatebox{270}{MARBERT} &\rotatebox{270}{QARiB} & \rotatebox{270}{mBERT} & \rotatebox{270}{AraELECTRA} & \rotatebox{270}{ALBERT} & \rotatebox{270}{XLM}   \\ \hline
\multirow{7}{*}{ATB} & 3.00\% & 0.914 & 0.915 & 0.929 & 0.916 & 0.924 & 0.924 & 0.868 & 0.935 & 0.938 \\
 & \cellcolor[HTML]{DAE8FC}{\color[HTML]{333333} 5.00\%} & \cellcolor[HTML]{DAE8FC}{\color[HTML]{333333} 0.934} & \cellcolor[HTML]{DAE8FC}{\color[HTML]{333333} 0.940} & \cellcolor[HTML]{DAE8FC}{\color[HTML]{333333} 0.949} & \cellcolor[HTML]{DAE8FC}{\color[HTML]{333333} 0.945} & \cellcolor[HTML]{DAE8FC}{\color[HTML]{333333} 0.950} & \cellcolor[HTML]{DAE8FC}{\color[HTML]{333333} 0.941} & \cellcolor[HTML]{DAE8FC}{\color[HTML]{333333} 0.912} & \cellcolor[HTML]{DAE8FC}{\color[HTML]{333333} 0.947} & \cellcolor[HTML]{DAE8FC}{\color[HTML]{333333} 0.953} \\
 & 7.00\% & 0.939 & 0.949 & 0.957 & 0.952 & 0.957 & 0.945 & 0.934 & 0.953 & 0.957 \\
 & 10.00\% & 0.943 & 0.953 & 0.960 & 0.957 & 0.961 & 0.947 & 0.945 & 0.954 & 0.960 \\
 & 20.00\% & 0.945 & 0.956 & 0.960 & 0.958 & 0.962 & 0.948 & 0.954 & 0.953 & 0.961 \\
 & 50.00\% & 0.940 & 0.953 & 0.955 & 0.954 & 0.958 & 0.941 & 0.957 & 0.948 & 0.955 \\
 & 100.00\% & 0.937 & 0.954 & 0.957 & 0.955 & 0.955 & 0.938 & 0.954 & 0.947 & 0.953 \\ \hline
 & 3.00\% & 0.769 & 0.767 & 0.803 & 0.784 & 0.787 & 0.725 & 0.658 & 0.763 & 0.714 \\
 & 5.00\% & 0.798 & 0.809 & 0.834 & 0.831 & 0.828 & 0.757 & 0.723 & 0.791 & 0.755 \\
 & 7.00\% & 0.791 & 0.811 & 0.842 & 0.840 & 0.845 & 0.755 & 0.789 & 0.782 & 0.761 \\
 & \cellcolor[HTML]{DAE8FC}10.00\% & \cellcolor[HTML]{DAE8FC}0.821 & \cellcolor[HTML]{DAE8FC}0.836 & \cellcolor[HTML]{DAE8FC}0.861 & \cellcolor[HTML]{DAE8FC}0.862 & \cellcolor[HTML]{DAE8FC}0.861 & \cellcolor[HTML]{DAE8FC}0.784 & \cellcolor[HTML]{DAE8FC}0.790 & \cellcolor[HTML]{DAE8FC}0.807 & \cellcolor[HTML]{DAE8FC}0.785 \\
 & 20.00\% & 0.822 & 0.844 & 0.868 & 0.864 & 0.866 & 0.796 & 0.824 & 0.809 & 0.797 \\
 & 50.00\% & 0.804 & 0.827 & 0.858 & 0.857 & 0.861 & 0.776 & 0.825 & 0.792 & 0.792 \\
\multirow{-7}{*}{CRS} & 100.00\% & 0.788 & 0.824 & 0.839 & 0.845 & 0.847 & 0.763 & 0.816 & 0.779 & 0.780 \\ \hline
 & 3.00\% & 0.753 & 0.780 & 0.798 & 0.766 & 0.783 & 0.732 & 0.683 & 0.779 & 0.753 \\
 & 5.00\% & 0.774 & 0.812 & 0.835 & 0.809 & 0.820 & 0.748 & 0.747 & 0.808 & 0.767 \\
 & 7.00\% & 0.788 & 0.831 & 0.847 & 0.830 & 0.834 & 0.757 & 0.776 & 0.815 & 0.783 \\
 & \cellcolor[HTML]{DAE8FC}10.00\% & \cellcolor[HTML]{DAE8FC}0.793 & \cellcolor[HTML]{DAE8FC}0.833 & \cellcolor[HTML]{DAE8FC}0.851 & \cellcolor[HTML]{DAE8FC}0.842 & \cellcolor[HTML]{DAE8FC}0.836 & \cellcolor[HTML]{DAE8FC}0.775 & \cellcolor[HTML]{DAE8FC}0.794 & \cellcolor[HTML]{DAE8FC}0.818 & \cellcolor[HTML]{DAE8FC}0.790 \\
 & 20.00\% & 0.794 & 0.840 & 0.857 & 0.850 & 0.851 & 0.768 & 0.809 & 0.814 & 0.806 \\
 & 50.00\% & 0.784 & 0.832 & 0.840 & 0.844 & 0.847 & 0.752 & 0.814 & 0.798 & 0.799 \\
\multirow{-7}{*}{DIA} & 100.00\% & 0.770 & 0.818 & 0.831 & 0.826 & 0.829 & 0.734 & 0.803 & 0.790 & 0.776 \\ \hline
 & 3.00\% & 0.829 & 0.876 & 0.879 & 0.879 & 0.885 & 0.809 & 0.840 & 0.864 & 0.833 \\
 & 5.00\% & 0.854 & 0.892 & 0.897 & 0.907 & 0.908 & 0.821 & 0.868 & 0.881 & 0.860 \\
 & \cellcolor[HTML]{DAE8FC}7.00\% & \cellcolor[HTML]{DAE8FC}0.860 & \cellcolor[HTML]{DAE8FC}0.901 & \cellcolor[HTML]{DAE8FC}0.910 & \cellcolor[HTML]{DAE8FC}0.916 & \cellcolor[HTML]{DAE8FC}0.917 & \cellcolor[HTML]{DAE8FC}0.832 & \cellcolor[HTML]{DAE8FC}0.882 & \cellcolor[HTML]{DAE8FC}0.885 & \cellcolor[HTML]{DAE8FC}0.865 \\
 & 10.00\% & 0.872 & 0.905 & 0.914 & 0.918 & 0.920 & 0.837 & 0.887 & 0.892 & 0.878 \\
 & 20.00\% & 0.880 & 0.908 & 0.917 & 0.922 & 0.923 & 0.846 & 0.900 & 0.893 & 0.878 \\
 & 50.00\% & 0.876 & 0.902 & 0.909 & 0.915 & 0.915 & 0.840 & 0.906 & 0.888 & 0.871 \\
\multirow{-7}{*}{DID} & 100.00\% & 0.864 & 0.892 & 0.896 & 0.903 & 0.903 & 0.823 & 0.906 & 0.877 & 0.858 \\ \hline
 & 3.00\% & 0.881 & 0.891 & 0.913 & 0.907 & 0.912 & 0.856 & 0.833 & 0.885 & 0.856 \\
 & \cellcolor[HTML]{DAE8FC}5.00\% & \cellcolor[HTML]{DAE8FC}0.899 & \cellcolor[HTML]{DAE8FC}0.910 & \cellcolor[HTML]{DAE8FC}0.926 & \cellcolor[HTML]{DAE8FC}0.923 & \cellcolor[HTML]{DAE8FC}0.924 & \cellcolor[HTML]{DAE8FC}0.876 & \cellcolor[HTML]{DAE8FC}0.876 & \cellcolor[HTML]{DAE8FC}0.898 & \cellcolor[HTML]{DAE8FC}0.880 \\
 & 7.00\% & 0.913 & 0.925 & 0.929 & 0.936 & 0.934 & 0.892 & 0.892 & 0.909 & 0.891 \\
 & 10.00\% & 0.908 & 0.920 & 0.931 & 0.930 & 0.929 & 0.890 & 0.901 & 0.905 & 0.897 \\
 & 20.00\% & 0.907 & 0.920 & 0.926 & 0.929 & 0.925 & 0.891 & 0.914 & 0.904 & 0.898 \\
 & 50.00\% & 0.899 & 0.909 & 0.918 & 0.919 & 0.915 & 0.876 & 0.911 & 0.889 & 0.884 \\
\multirow{-7}{*}{GMR} & 100.00\% & 0.890 & 0.900 & 0.910 & 0.909 & 0.908 & 0.865 & 0.901 & 0.880 & 0.878 \\ \hline
\end{tabular}
}
\caption{Performance per models using different threshold $\delta$}
    \label{tab:thresholdselection}

\end{table*}

\begin{figure*}[ht]
\begin{subfigure}{0.33\linewidth}
  \centering
  \includegraphics[width=\linewidth]{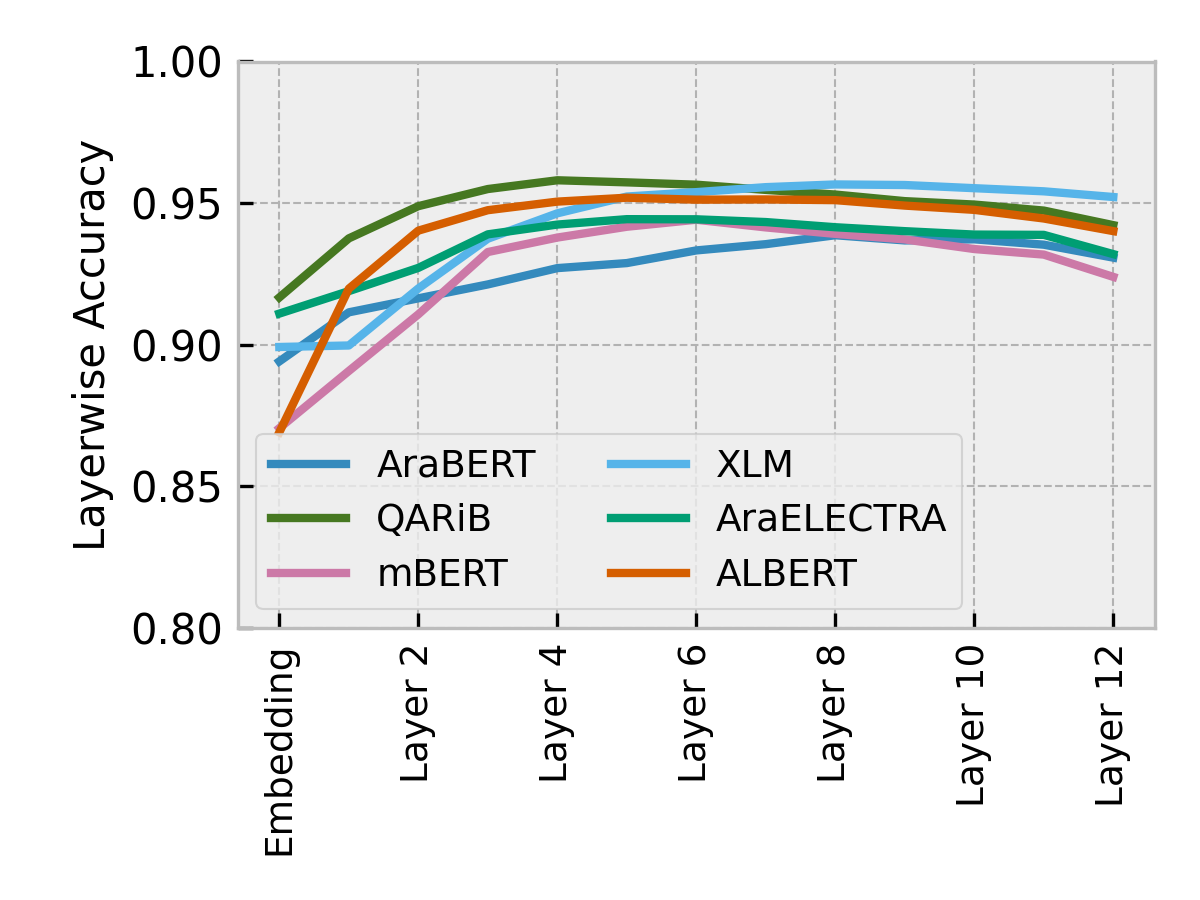}  
  \caption{ATB}
  \label{fig:atb_layerwise_acc}
\end{subfigure}
\begin{subfigure}{0.33\textwidth}
  \centering
  \includegraphics[width=\linewidth]{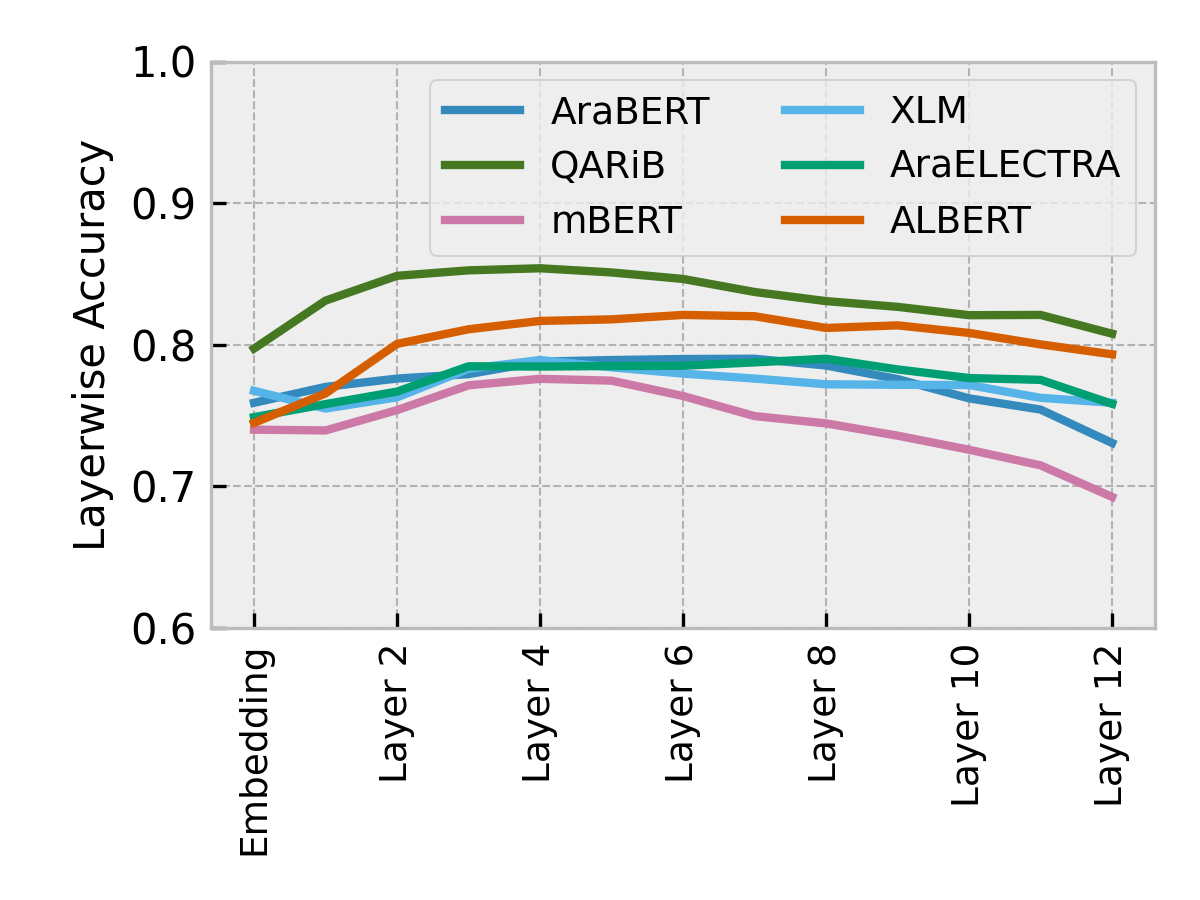}
  \caption{DIA}
  \label{fig:dia_layerwise_acc}
\end{subfigure} 
\begin{subfigure}{0.33\textwidth}
  \centering
  \includegraphics[width=\linewidth]{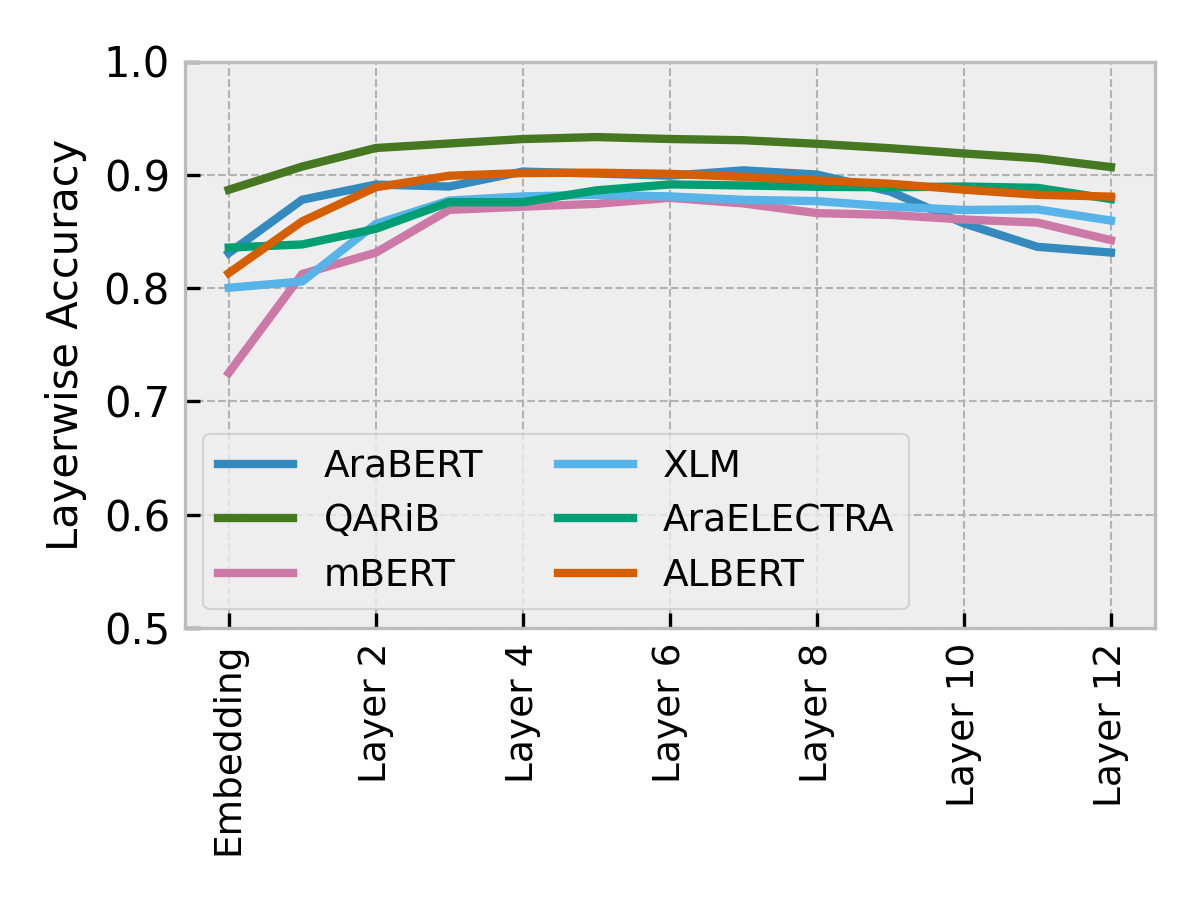}
  \caption{GMR}
  \label{fig:gmr_layerwise_acc}
\end{subfigure}
\caption{Layer-wise accuracy for ATB, DIA, GMR tasks.}
\label{fig:layers_accuracy_apdx}
\end{figure*}

\begin{figure*}[ht]
\centering
	\begin{subfigure}{0.7\textwidth}
		\centering
		\includegraphics[trim=0 0 0 0,clip,width=\linewidth]{figures/models_neurons_per_property_ATB.png}
		\caption{ATB}
		\label{fig:neurons_per_prop-atb-a}
	\end{subfigure}
	\begin{subfigure}{0.5\textwidth}
		\centering
		\includegraphics[trim=0 0 0 0,clip,width=\linewidth]{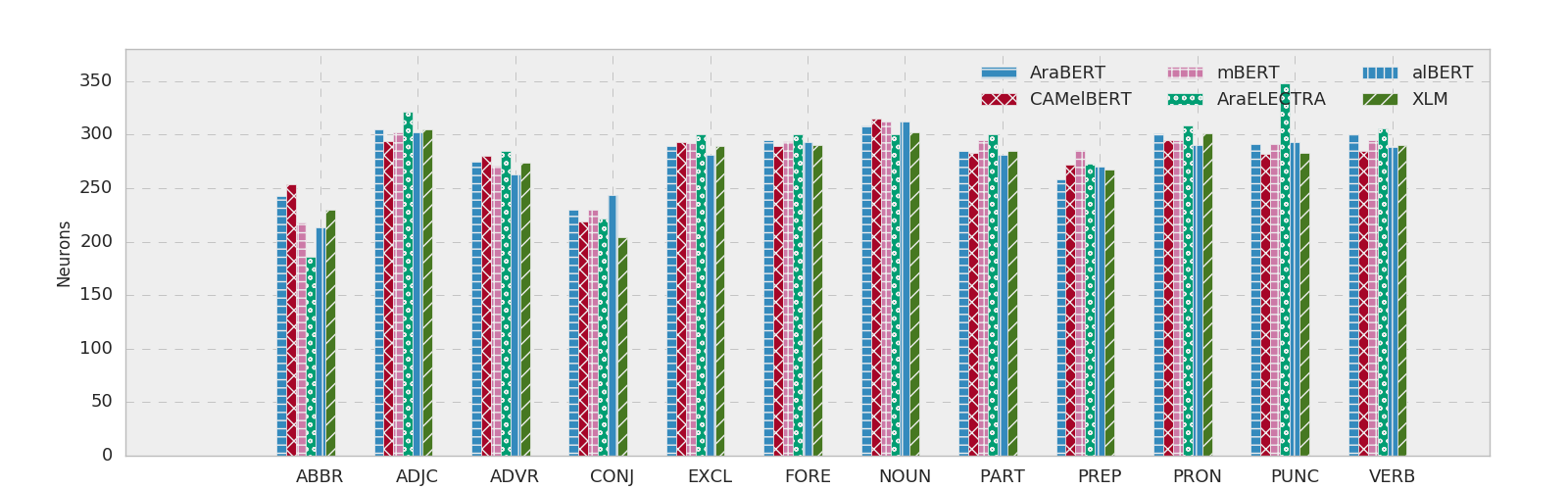}
		\caption{CRS}
		\label{fig:neurons_per_prop-crs-a}
	\end{subfigure}
	\begin{subfigure}{0.5\textwidth}
		\centering
		\includegraphics[width=\linewidth]{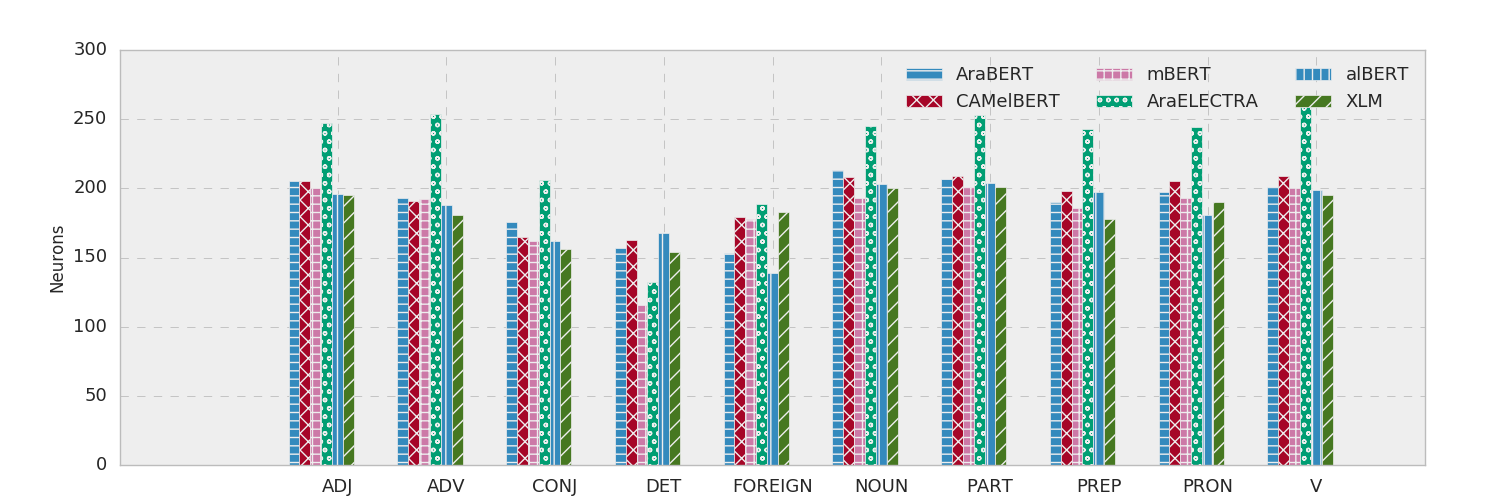}  
		\caption{DIA}
		\label{fig:neurons_per_prop-dia-a}
	\end{subfigure}
	\begin{subfigure}{0.3\textwidth}
		\includegraphics[width=\linewidth]{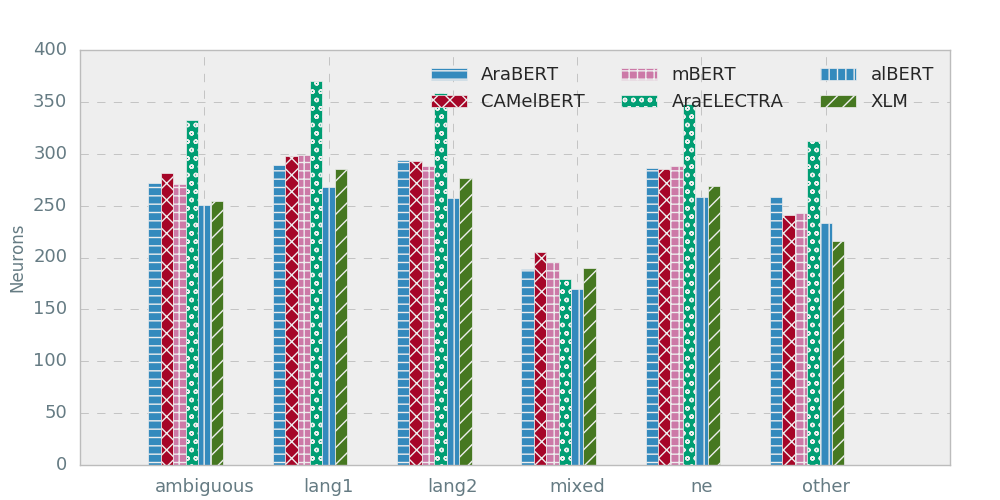}  
		\caption{DID}
		\label{fig:neurons_per_prop-did-a}
	\end{subfigure}

	\begin{subfigure}{0.7\textwidth}
		\centering
		\includegraphics[trim=0 0 0 0,clip,width=\linewidth]{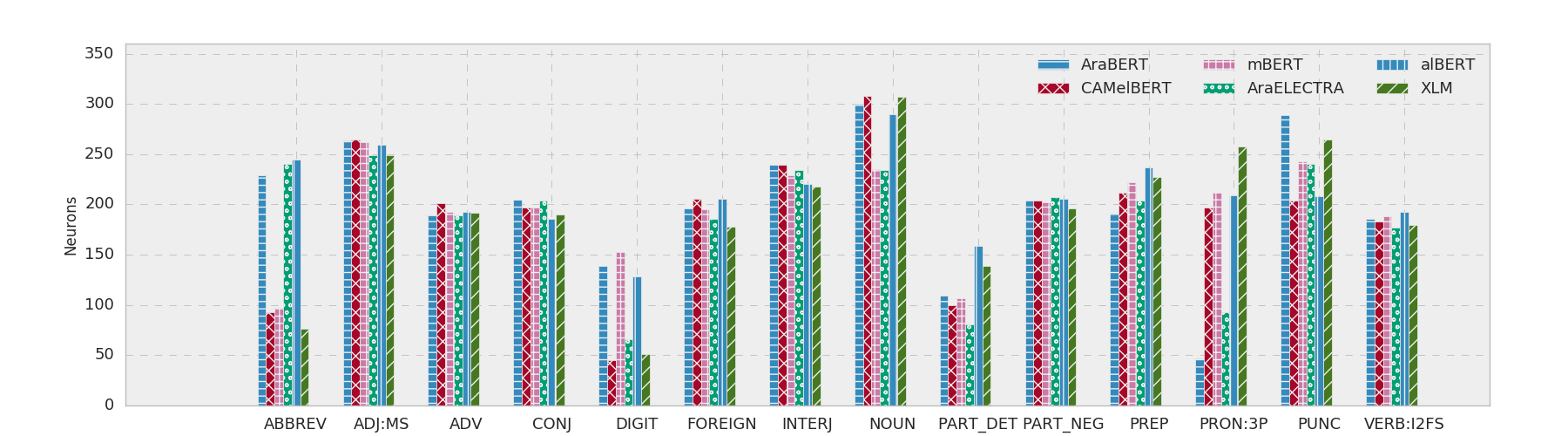}
		\caption{GMR}
		\label{fig:neurons_per_prop-gmr-a}
	\end{subfigure}
	\squeezeupless
	\caption{Distribution of neurons per property}
	\label{fig:neurons_per_prop_appdx}
\end{figure*}

\begin{figure*}[ht]
\begin{subfigure}{0.32\textwidth}
  \centering
  \includegraphics[width=\linewidth]{figures/new-atb-layerwise-selection.png}
  \caption{ATB}
  \label{fig:atb_neurons_layer}
\end{subfigure}
\begin{subfigure}{0.32\textwidth}
  \centering
  \includegraphics[width=\linewidth]{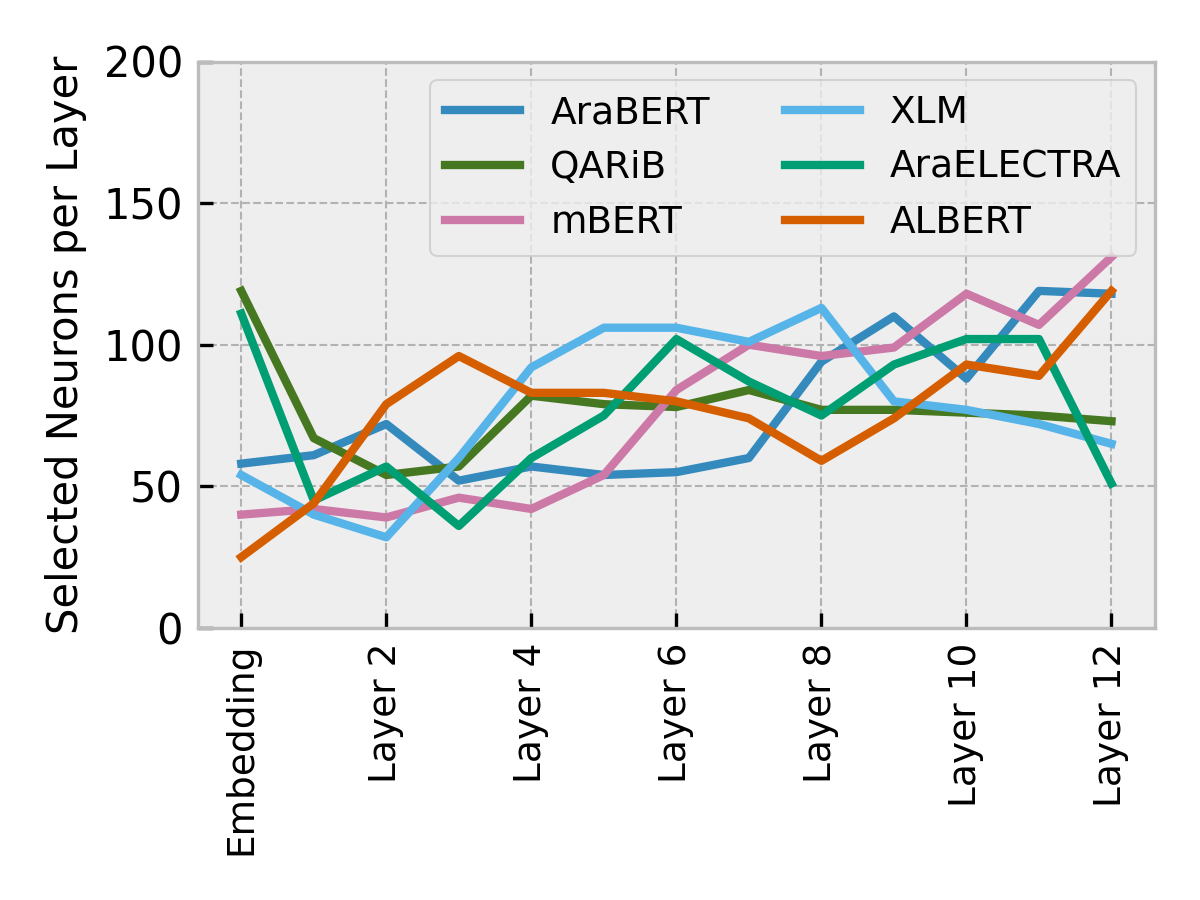}
  \caption{CRS}
  \label{fig:dia_neurons_layer}
\end{subfigure}
\begin{subfigure}{0.32\textwidth}
  \centering
  \includegraphics[width=\linewidth]{figures/new-dia-layerwise-selection.png}
  \caption{DIA}
  \label{fig:did_neurons_layer}
\end{subfigure}
\centering
\begin{subfigure}{0.32\textwidth}
  \centering
  \includegraphics[width=\linewidth]{figures/new-did-layerwise-selection.png}
  \caption{DID}
  \label{fig:atb_neurons_layer}
\end{subfigure}
\begin{subfigure}{0.32\textwidth}
  \centering
  \includegraphics[width=\linewidth]{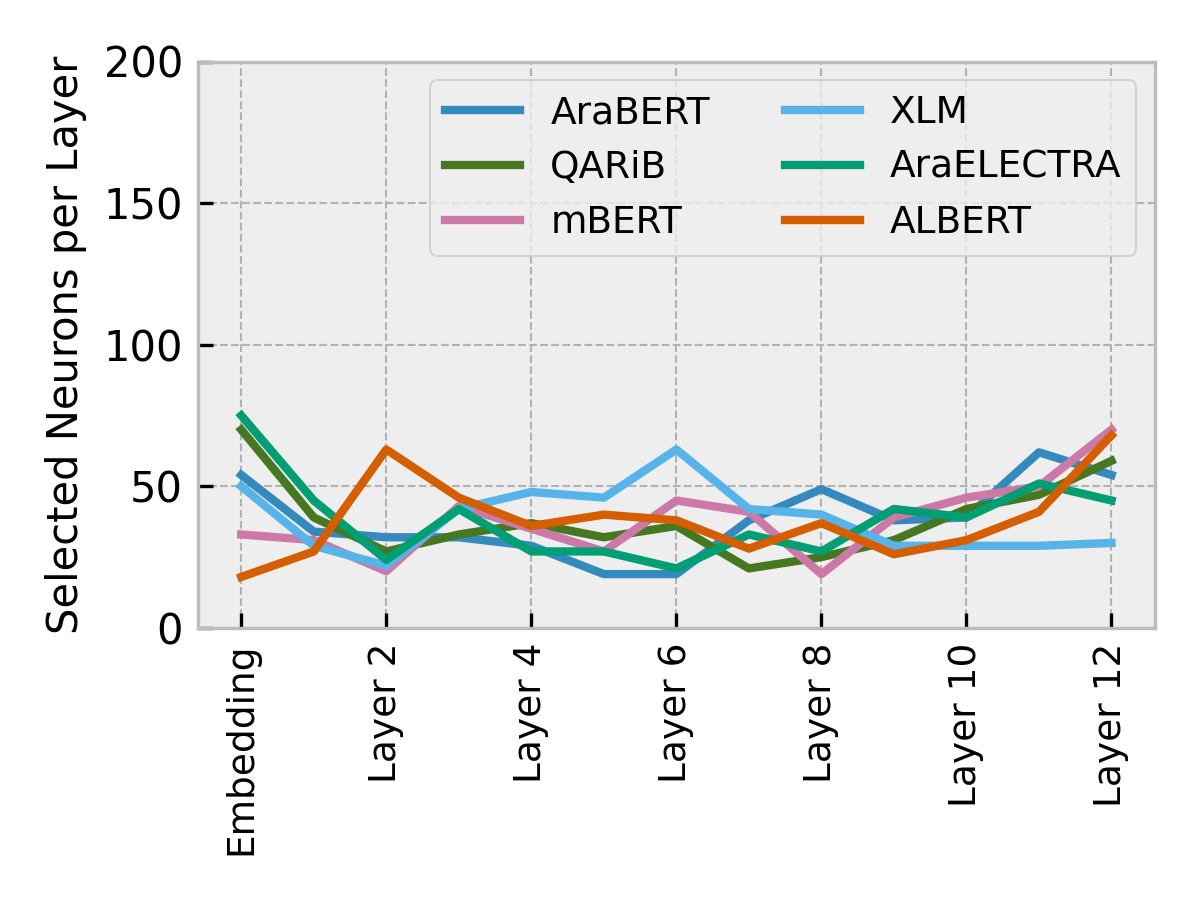}
  \caption{GMR}
  \label{fig:did_neurons_layer}
\end{subfigure}
\caption{Distribution of selected neurons across the layers}
\label{fig:neurons_layer_app}
\end{figure*}

\endgroup

\end{document}